\documentclass[a4paper,fleqn]{cas-dc}

\usepackage[numbers]{natbib}
\usepackage[ruled,linesnumbered]{algorithm2e}  
\usepackage{algorithm2e}
\SetKwInput{KwIn}{Input}
\SetKwInput{KwOut}{Output}
\SetKwInput{KwData}{Data}
\SetKwInput{KwResult}{Result}
\SetKwFunction{BuildGraph}{BuildTemporalGraph}
\SetKwFunction{Extract}{ExtractTrajectories}
\SetKwFunction{Eval}{EvaluateTrajectories}
\SetKwRepeat{Do}{do}{while}

\usepackage{booktabs}
\usepackage{multirow}
\usepackage{array}
\usepackage{ragged2e}
\usepackage{makecell} 

\def\tsc#1{\csdef{#1}{\textsc{\lowercase{#1}}\xspace}}
\tsc{WGM}
\tsc{QE}
\tsc{EP}
\tsc{PMS}
\tsc{BEC}
\tsc{DE}

\begin{document}

\let\WriteBookmarks\relax
\def\floatpagepagefraction{1}
\def\textpagefraction{.001}

\shorttitle {Temporal Point-Supervised Signal Reconstruction: A Human-Annotation-Free Framework for Weak Moving Target Detection}
\shortauthors{Weihua Gao et~al.}

\title [mode = title]{Temporal Point-Supervised Signal Reconstruction: A Human-Annotation-Free Framework for Weak Moving Target Detection}                      
\tnotemark[1,2]

\tnotetext[1]{This work was supported in part by the Youth Innovation Promotion Association under Grant E1213A02; and in part by the Key Research Program of Frontier Sciences, Chinese Academy of Sciences (CAS), under Grant 22E0223301.}
\cortext[cor1]{Corresponding author}

\author[1,2]{Weihua Gao}[style=chinese]
\ead{gaoweihua22@mails.ucas.ac.cn}

\author[1]{Chunxu Ren}[style=chinese]
\ead{renchunxu23@mails.ucas.ac.cn}

\author[1]{Wenlong Niu}[style=chinese]
\cormark[1]
\ead{niuwenlong@nssc.ac.cn}

\author[1]{Xiaodong Peng}[style=chinese]
\ead{pxd@nssc.ac.cn}

\affiliation[1]{organization={Key Laboratory of Electronics and Information Technology for Space Systems, National Space Science Center, Chinese Academy of Sciences},
            city={Beijing},
            postcode={100190},
            country={China}}

\affiliation[2]{organization={School of Computer Science and Technology, University of Chinese Academy of Sciences},
            city={Beijing},
            postcode={100049},
            country={China}}
\begin{abstract}
In low-altitude surveillance and early warning systems, detecting weak moving targets remains a significant challenge due to low signal energy, small spatial extent, and complex background clutter. Existing methods struggle with extracting robust features and suffer from the lack of reliable annotations.
To address these limitations, we propose a novel Temporal Point-Supervised  (TPS) framework that enables high-performance detection of weak targets without any manual annotations. 
Instead of conventional frame-based detection, our framework reformulates the task as a pixel-wise temporal signal modeling problem, where weak targets manifest as short-duration pulse-like responses. A Temporal Signal Reconstruction Network (TSRNet) is developed under the TPS paradigm to reconstruct these transient signals.
TSRNet adopts an encoder–decoder architecture and integrates a Dynamic Multi-Scale Attention (DMSAttention) module to enhance its sensitivity to diverse temporal patterns. Additionally, a graph-based trajectory mining strategy is employed to suppress false alarms and ensure temporal consistency.
Extensive experiments on a purpose-built low-SNR dataset demonstrate that our framework outperforms state-of-the-art methods while requiring no human annotations. It achieves strong detection performance and operates at over 1000 FPS, underscoring its potential for real-time deployment in practical scenarios.
\end{abstract}

\begin{graphicalabstract}
\includegraphics[scale=0.3]{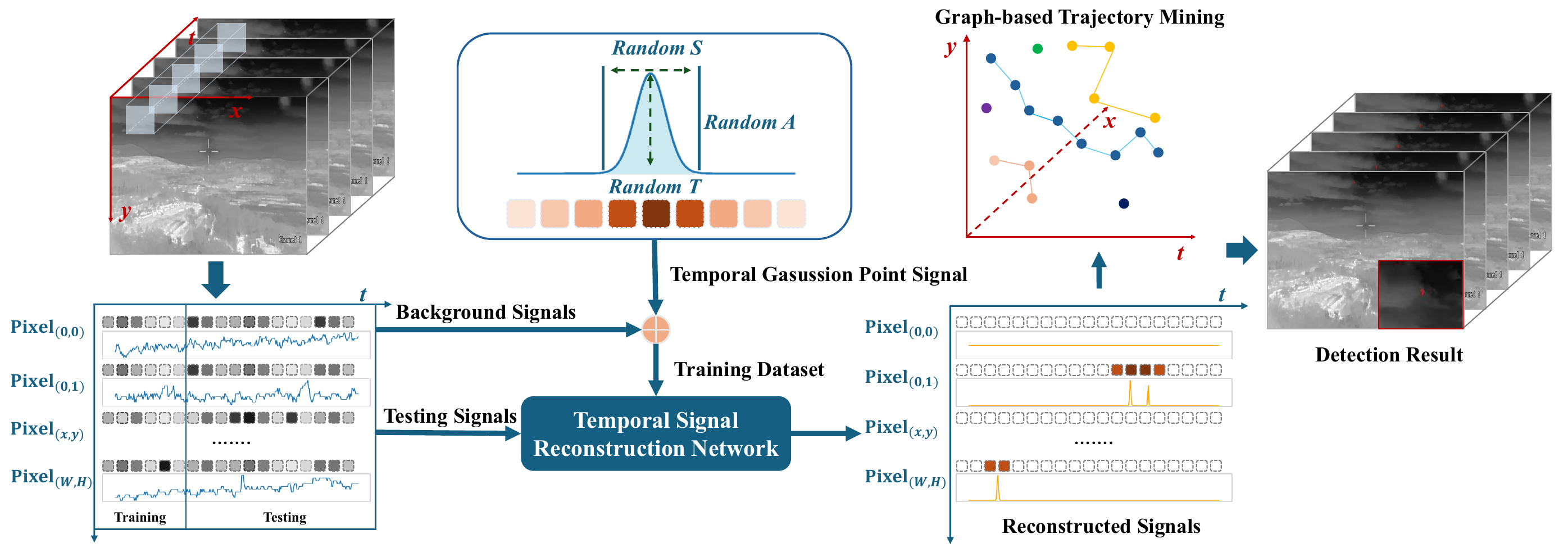}
\end{graphicalabstract}

\begin{highlights}
\item Eliminates human annotations for low-SNR target detection
\item Extremely fast detection speed under extremely low computational load, suitable for practical deployment
\item The performance on more challenging low-SNR dataset surpasses existing fully supervised methods
\end{highlights}


\maketitle

\section{Introduction}
In typical scenarios such as aerial surveillance, low-altitude monitoring, and remote sensing observation, small moving targets—such as drones or flying birds—are often the primary targets of interest. However, due to their long distance, small size, and low contrast, these targets usually occupy only a sub-pixel to a few pixels on the imaging plane. As a result, they lack clear edge, texture, or shape information and are easily overwhelmed by complex background clutter and sensor noise, leading to extremely low signal-to-noise ratios (SNR) ~\cite{Chen_2024_SSTNeta, ZHU2024109362}. Such "weak and small targets" differ significantly from the recognizable objects in general visual tasks, and have long been considered one of the most challenging problems in computer vision.

In recent years, researchers have increasingly turned their attention to the more challenging yet highly valuable task of weak target detection ~\cite{10772282}. Existing weak target detection methods can be broadly categorized into two groups: model-driven approaches, which construct rule-based detectors using handcrafted priors such as contrast, saliency, or background modeling to capture spatiotemporal differences between targets and background ~\cite{10757439, 10646209, 9979804, 9274351}; and learning-driven approaches, which rely on neural networks to automatically learn expressive features and perform target detection in a data-driven manner ~\cite{Tong_2024_ST_Trans, Xiao_2022_DSFNet, 10606537, 9165022}.

In model-driven approaches, researchers typically enhance weak and small targets based on specific prior features. These methods can be broadly categorized into three groups: filtering-based enhancement techniques, contrast enhancement methods inspired by the human visual system, and low-rank and sparse modeling approaches for background suppression.
Filtering-based methods aim to enhance the contrast between targets and background by designing local operators. Representative techniques include Max-Mean, Top-Hat filters, wavelet transforms, and Laplacian of Gaussian (LoG) enhancement, which have been widely adopted in infrared small target detection tasks ~\cite{10757439,9979804,Zeng_2006_design,Deshpande_1999}.
HVS-inspired methods simulate the human eye’s sensitivity to local contrast and motion saliency. These approaches typically boost target responses through local contrast enhancement, directional feature extraction, and motion-aware mechanisms ~\cite{10646209,STLDM,chen2013local,chen2023simplified,9130832}.
Low-rank and sparse decomposition methods model image sequences as the superposition of a low-rank background and sparse targets. Techniques based on Robust Principal Component Analysis (RPCA) or matrix factorization are often employed to remove the background and extract salient targets ~\cite{li2023sparse,liu2023single,PSTNN,sun2018infrared,gao2013infrared}. However, these methods often rely on strong prior assumptions—such as the target being a bright spot or the background being static—which may not hold in real-world scenarios. Consequently, their performance tends to degrade in complex dynamic backgrounds or low SNR conditions, where robustness and adaptability become critical.

Compared to model-driven approaches, learning-based methods leverage neural networks to automatically extract representative features of targets, significantly improving the performance of weak and small target detection. Based on the types of features they exploit, these methods can be broadly categorized into appearance-based approaches and those that integrate spatiotemporal features.

Most of the early studies were based on single-frame detection methods made suitable for small target scenes by combining a priori knowledge, resolution enhancement strategies or multi-scale fusion mechanisms~\cite{LI2023108962,LIN2022103684}.
However, due to the inherently vague and semantically poor nature of weak targets, approaches relying solely on appearance modeling often face performance bottlenecks and struggle under complex background interference.

To overcome this limitation, recent research has increasingly focused on temporal modeling. By introducing temporal windows or stacking sequential frames, these methods aim to capture the continuous motion patterns of targets, thereby enhancing both robustness and detection accuracy ~\cite{Xiao_2022_DSFNet,Tong_2024_ST_Trans,Huang_2024_LMAFormer,Chen_2024_SSTNeta}. Through the integration of spatiotemporal behavioral cues, these approaches have shown promising potential in mitigating the limitations caused by weak appearance signals.

However, these methods still face several critical limitations. First, they rely heavily on high-quality annotated data. In weak target scenarios, the targets are often faint and have blurred boundaries, making precise annotation a challenging and costly task. Second, most existing studies only use temporal cues as a supplement to appearance features, which cannot cope with extremely weak targets. Finally, the high computational requirements and inference latency of deep models limit their applicability in scenarios requiring real-time performance or deployment efficiency.

To alleviate the high cost of annotation, a number of recent works have explored weakly-supervised and semi-supervised approaches for small target detection ~\cite{Cui_2025_Weakly,Kou_2024_MCGCa,Li_2023_Monte,Ni_2025_Point-to-Point,Ying_2023_Mapping,Yuan_2025_Full}. Traditional annotation typically involves pixel-level masks or bounding boxes, known as full supervision. In contrast, point-supervised methods require only a single point per target, aiming to approximate fully supervised performance while significantly reducing labeling effort. While these approaches do reduce annotation overhead to some extent, they still fundamentally depend on manual labeling. Moreover, representing an entire target with only one point inevitably introduces label bias, which can negatively impact detection accuracy.

Therefore, fundamentally reducing the reliance on high-quality manual annotations and exploring more autonomous and generalizable detection strategies has become a critical challenge in applying weak target detection to real-world scenarios. To this end, it is necessary to revisit the essence of moving target detection, which fundamentally aims to determine the target’s position in every frame—that is, to acquire its full $(x,y,t)$ information. Given the difficulty of localizing weak targets in the image frame (i.e., identifying $(x,y)$ coordinates), we propose an alternative perspective that shifts the detection unit from frame-wise imagery to pixel-wise temporal signals. If we can determine when a target appears at each pixel, estimate the temporal component $t$—we can still achieve effective target detection.

Motivated by this insight, we propose a novel Temporal Point-Supervised  (TPS) framework for weak moving targets, aiming to eliminate dependence on manual annotations by approaching the problem from the perspective of pixel-level temporal signal reconstruction. Specifically, we observe that although weak targets may appear vague in an image, they often exhibit brief, pulse-like responses along the temporal dimension. By modeling and recovering such transient patterns, it becomes possible to detect targets using only temporal point supervision. To ensure generalizability, we avoid imposing rigid assumptions on the pulse shape and instead model target signals using a Gaussian probability distribution. By synthesizing target-like pulses with varying intensities and scales into background pixel sequences, we establish a data generation mechanism that drives the network to learn dynamic target representations.

In implementation, we design a Temporal Signal Reconstruction Network (TSRNet) based on an encoder–decoder architecture, enhanced with a Dynamic Multi-Scale Attention  (DMSAttention) mechanism to improve responsiveness to diverse target patterns. To address the severe class imbalance caused by extreme signal sparsity, we introduce a weighted loss function, which improves training stability and reconstruction accuracy. Although this signal reconstruction stage already performs well in suppressing background interference, under extremely low SNR conditions, some background fluctuations may still be mistakenly identified as target signals. To further reduce false positives, we incorporate a Graph-based Trajectory Mining (GTM) algorithm that aggregates and filters detection points by exploiting spatiotemporal continuity. Additionally, to mitigate the impact of parameter sensitivity in the trajectory extraction process, we employ a Monte Carlo optimization strategy for automatic parameter tuning.

Ultimately, the proposed framework achieves continuous detection and robust trajectory recovery of weak moving targets without relying on any manual annotations. The main contributions of this work are summarized as follows:
\begin{enumerate}
    \item We propose a high-performance TPS-based supervised framework for weak moving target detection. By synthesizing temporal point signals into background pixels to generate training data, the framework completely eliminates the need for human-labeled supervision, offering strong generalization and practical applicability.
    \item We reformulate the detection paradigm from frame-based target detection to pixel-wise temporal signal modeling. From this novel perspective, we design an encoder–decoder-based temporal reconstruction network enhanced with dynamic multi-scale attention mechanisms, enabling effective recovery of weak target signals while suppressing background interference.
    \item To address the challenges of extreme signal sparsity and severe class imbalance, we design a weighted loss function that significantly enhances training stability and improves signal reconstruction quality. In addition, we introduce a graph-based trajectory mining algorithm, combined with a Monte Carlo optimization strategy, to achieve unsupervised false alarm suppression.
    \item In the more challenging low SNR dataset we constructed, our framework achieved best performance against other state-of-the-art methods with an average FPS of 1000+.
\end{enumerate}

\section{Related Work}
This section reviews learning-based weak target detection methods, categorized by their supervision strategies.
\subsection{Fully Supervised Detection Methods}
Currently, the vast majority of weak target detection algorithms are fully supervised. Given the abundance of fully supervised weak target detection approaches, we organize the review based on the types of features extracted: appearance-based methods and appearance-motion joint feature methods.
\subsubsection{Appearance-Based Detection Methods}

Appearance-based detection methods typically take single-frame as input, using convolutional neural networks (CNNs) or other deep learning models to enhance the spatial appearance features of weak targets, thereby improving SNR and detection performance. These methods often integrate classical techniques such as contrast enhancement, pyramid structures, and attention mechanisms within the network design.

Early representative work includes Bruce McIntosh et al. ~\cite{9199539}, who pioneered learning-based weak target detection by enhancing spatial contrast of targets in infrared images using CNNs to improve signal-to-clutter ratio. Dai et al. ~\cite{9314219} embedded local contrast priors into their network structure, achieving detection performance surpassing traditional model-driven methods on the NUDT-SIRST dataset. Subsequent research focused on refining feature extraction methods: Lian Huang et al. ~\cite{HUANG2021103755} designed a multi-scale pyramid structure combined with an attention fusion mechanism to strengthen weak target feature representation; Moran Ju et al. ~\cite{JU2021103659} incorporated image filtering modules to enhance target regions and suppress background interference.

To better address the scale characteristics of small target detection, Lianghui Ding et al. ~\cite{DING2021102949} introduced high-resolution feature layers and adaptive pipeline filters into the SSD framework to correct detection errors; Yimian Dai et al. ~\cite{Dai_2021_WACV} proposed an asymmetric context modulation module that interacts global context feedback with low-level attention channels to boost response to faint targets. Some works have also incorporated global modeling mechanisms such as Transformers—for example, Jian Lin et al. ~\cite{LIN2022103684} proposed the U-Transformer network to capture long-range background dependencies, thereby more effectively suppressing clutter.

Furthermore, to mitigate the loss of small target features in deep networks, Boyang Li et al. ~\cite{li2022dense} constructed dense nested interaction modules for multi-scale feature fusion; Shanliang Liu et al. ~\cite{10109684} employed similarity enhancement mechanisms to dynamically adjust target weights and introduced parallel branches to improve robustness. Xin Wu et al. embedded a small U-Net within a larger U-Net backbone to enable multi-level, multi-scale representation learning, enhancing both global and local contrast ~\cite{wu2022uiu}. Longyuan Guo et al. ~\cite{10766378} improved residual structures and integrated multi-scale pyramids with attention mechanisms to effectively alleviate edge feature loss caused by information skipping. Tianfang Zhang et al. ~\cite{10024907} utilized an attention-guided context-aware module combining local semantics and global relational modeling to optimize detection accuracy. Qiang Li et al. ~\cite{10830282} proposed MMLNet with edge, localization, and detection branches that dynamically integrate multi-scale information via mutual guidance modules to enhance small target completeness and accuracy. Xiaoyu Xu et al. ~\cite{10772282} adopted sparse sampling and hybrid filtering techniques to capture complex shapes and edge information of weak targets, supplemented by multi-scale feature enhancement modules emphasizing key small target characteristics.

Despite these advances, the aforementioned methods heavily rely on spatial appearance features, which are challenging to extract effectively under extremely low SNR conditions. Moreover, they largely overlook the temporal continuity of targets and fail to fully exploit dynamic evolution across multiple frames. These methods still depend on precise or approximate label supervision and struggle to adapt to practical scenarios where targets are sparse and their locations are difficult to annotate.

\subsubsection{Detection Methods Based on Joint Appearance-Motion Features}
Due to the scarcity of reliable appearance information in weak targets, appearance-based detection methods often encounter performance bottlenecks in practical applications. To overcome this limitation, researchers have gradually incorporated motion information as a complementary cue, leveraging the temporal continuity of targets to build more robust spatiotemporal feature representations. Detection methods based on joint appearance-motion modeling typically take consecutive frame sequences as input and employ sophisticated spatiotemporal neural networks to extract dynamic behavioral patterns of targets, thereby enhancing the detectability of weak signals.

For example, Du et al.~\cite{9570298} proposed a frame-to-frame energy accumulation enhancement mechanism that significantly amplifies target responses in temporal image sequences and achieves strong performance in cluttered infrared scenarios. Chen et al.~\cite{Chen_2024_SSTNeta} exploited cross-slice ConvLSTM nodes to capture spatiotemporal structures, introducing motion coordination loss and a coupled neck design to effectively fuse multi-frame information, setting new state-of-the-art results across multiple public datasets. Huang et al.~\cite{Huang_2024_LMAFormer} designed a local motion-aware spatiotemporal attention mechanism combined with a multi-scale Transformer encoder to jointly model local target motion and dynamic background, effectively mitigating interference from dynamic clutter.

In terms of network architecture, Li et al.~\cite{li2025dbmstn} introduced a dual-branch multi-scale spatiotemporal network (DBMSTN) integrating saliency enhancement, multi-level fusion, and weighted loss to address scale variation and class imbalance challenges in small target detection. Peng et al.~\cite{peng2025moving} focused on efficient spatiotemporal modeling by proposing a lightweight dual-branch fusion network that adaptively adjusts the fusion of full-video temporal features and keyframe spatial features through a self-optimization module, reducing computational cost. S. Zhu et al.~\cite{zhu2024tmp} emphasized the complementarity of spatial and temporal modeling by designing spatial auxiliary and motion feature branches for collaborative fusion, employing a symmetric weighting module for multidimensional feature integration. Tong et al.~\cite{Tong_2024_ST_Trans} further introduced Transformer architectures combined with spatiotemporal transformation modules (STTM) to capture cross-frame dependencies, improving detection accuracy in complex dynamic scenes.

Despite significant performance gains via spatiotemporal fusion, these methods generally suffer from complex architectures, high computational overhead, and strong reliance on labeled data. Furthermore, they still struggle to precisely capture the weak responses of extremely low-SNR, sparse transient targets, resulting in limited robustness.

\subsection{Point-Supervised and Weakly Supervised Detection Methods}

Although fully supervised detection methods achieve excellent performance, they rely heavily on precise annotations. As noted earlier, obtaining accurate labels for weak target detection tasks is costly and prone to bias and errors. Inspired by the lightweight supervision paradigm in general object detection, recent research has explored weak supervision approaches using single-point labels or pseudo-labels as substitutes for precise annotations, yielding encouraging preliminary results.

In image segmentation, Bearman et al. ~\cite{bearman2016s} first proposed point supervision, demonstrating that under reasonable guidance mechanisms, single-point annotations can replace precise masks for model training. This idea has since been introduced into weak small target detection. Ying et al. ~\cite{Ying_2023_Mapping} found that neural networks initially learn pixel clusters around target points and gradually converge to precise locations, enabling self-recovery of target masks from point-only labels. Li et al. ~\cite{Li_2023_Monte} introduced random perturbations by adding noise, followed by multiple clustering and averaging to generate stable pseudo-masks, effectively converting any fully supervised detection network into a point-supervised weak detector. Kou et al. ~\cite{Kou_2024_MCGCa} proposed a multi-scale chain-growth clustering algorithm that utilizes Euclidean decay and neighborhood rules to automatically generate reliable pseudo-labels from point annotations. Yuan et al. ~\cite{Yuan_2025_Full} treated point-guided mask generation as a small target detection task with positional prompts, designing a full pipeline from energy initialization, dual-prompt embedding, to bounding box matching. Ni et al. ~\cite{Ni_2025_Point-to-Point} directly modeled detection as a point regression problem, achieving weak target localization via high-resolution outputs. Cui et al. ~\cite{Cui_2025_Weakly} generated coarse pseudo-masks through a teacher network with limited pixel labels and combined random walk and optimization modules to refine pseudo-label quality, subsequently enhancing weakly supervised detection performance with a student network.

Besides point supervision, some researchers have explored weak supervision using pseudo-labels and synthetic data. Lu et al. ~\cite{10496142} developed an infrared sequence noise modeling framework and negative augmentation strategy to expand training data at the data level, aiding network learning of target features. Duan et al. ~\cite{10824834} constructed a teacher-student framework that jointly trains the student network with limited labeled data and abundant pseudo-labels, achieving performance close to or even surpassing fully supervised methods.

While these approaches alleviate label dependence, they still rely on supervision, often require pseudo-mask generation, pseudo-label fitting, or multi-stage training, resulting in complex training pipelines and uncertain pseudo-label quality.

\section{Proposed Method}
\begin{figure*}[!t]
	\centering
	\includegraphics[width=\linewidth]{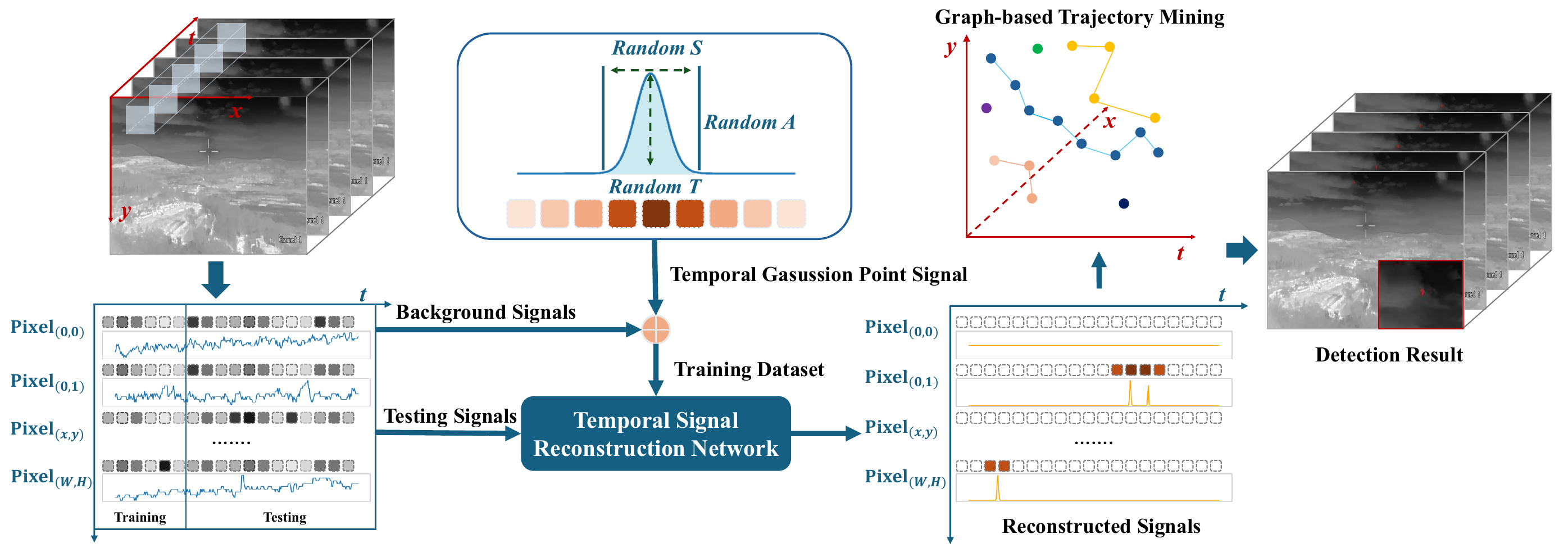}
	\caption{Overview of the proposed TPS-based framework for detecting weak moving targets. The framework begins with the extraction of temporal intensity signals from each pixel in the input image sequence. During training, sparse temporal Gaussian point signals are injected as supervision to indicate target presence. TSRNet is trained to selectively enhance target signals while suppressing background and noise. In the inference stage, the trained TSRNet processes unseen temporal signals to generate reconstructed signals. Finally, a graph-based trajectory mining algorithm is applied to the reconstructed signals to extract coherent target trajectories.}
	\label{framework_fig}
\end{figure*}

Figure~\ref{framework_fig} illustrates the overall architecture of our proposed TPS-based framework, which consists of two stages: training and testing. A key premise of our approach is that a dedicated model is trained for each scene, enabling the network to learn dynamic characteristics specific to that environment.

During the training stage, we extract the pixel-level temporal background signals from an image sequence. Whether real targets are present in this sequence is not critical—what matters is the background signal feature it provides. We then superimpose temporal point signals—modeled as 1D Gaussian heat maps—onto these background signals. The pulse parameters, including amplitude $A$, onset time $T$, and duration 
$S$, are randomly sampled to simulate diverse target behaviors. These composite signals serve as training samples for the TSRNet, which learns to recover potential target pulses embedded within complex backgrounds.

In the testing stage, image sequences from the same scene are again converted into pixel-wise temporal signals and passed through the trained network for reconstruction. The TSRNet output significantly suppresses background interference while enhancing target-relevant signals. By extracting salient points from the reconstructed signal, we obtain candidate target positions in $(x,y,t)$ space. Finally, a graph-based trajectory mining algorithm is applied to link sparse detections over time, yielding complete motion trajectories and final detection results.

\subsection{Temporal Signal Modeling at the Pixel Level}
\subsubsection{Motivation for the Temporal Point Supervision Mechanism}
The essential objective of target detection is to obtain the complete spatiotemporal information of a target, represented as a triplet $(x, y, t)$. Most existing methods adopt a frame-based paradigm, where the spatial location $(x, y)$ is inferred from the $t$-th frame. However, in the case of weak and small target detection, the targets often exhibit blurred edges, extremely small sizes, and very low SNR. These characteristics make it extremely difficult to accurately localize the target in image frames, posing significant challenges for traditional detection algorithms and introducing considerable subjectivity and uncertainty in manual annotations.

To address this challenge, we reformulate the detection task as a temporal signal modeling problem based on pixel-wise intensity variations over time. Specifically, we treat each pixel $(x, y)$ as a unit and extract its corresponding temporal intensity signal. We then determine whether a transient pulse—induced by a passing target—exists in the pixel's signal, thereby inferring the target’s time of occurrence $t$. Compared to spatial localization in complex image backgrounds, this modeling approach offers significant advantages: background signals typically exhibit slow variations over time (e.g., sky, clouds), whereas the movement of a target introduces sharp, short-duration perturbations ~\cite{Wang_2023_Dim}, making such temporal features more conducive to effective modeling and detection.

More importantly, this temporal modelling approach provides a viable way to discard manual annotation. Since it eliminates the need for spatial annotations, the network can be trained using only point-labels constructed from temporal signals. In theory, it suffices to supervise the time point $t$ when a target appears at a given pixel—an idea conceptually similar to “single point supervision” in frame-based detection. However, real-world targets often exhibit diffusion effects, with their influence spanning multiple frames. Relying solely on binary labels at a single time frame may lead to unstable training and blurred target representations.

To address this issue, we draw inspiration from the Gaussian heatmap strategy commonly used in center-based object detection ~\cite{Zhou_2019_Objects, Ni_2025_Point-to-Point}, and design the temporal Gaussion heatmap as supervision. This temporal Gaussion point signal peaks at the target’s central time point $T$ and decays smoothly as the time deviates from $T$. By randomly sampling the amplitude $A$, center $T$, and width $S$ of the signal, we can flexibly simulate targets with varying speeds, durations, and appearances. We refer to this strategy as the Temporal Point Supervision (TPS) mechanism.

What makes this mechanism particularly compelling is its strong physical consistency with the Point Spread Function of real targets. Due to the diffraction limit of optical systems, a point source target typically appears as a blurred spot in the image, which can be modeled by a two-dimensional Gaussian function ~\cite{10681464}, as shown in Equation~\ref{PSF_equ}:

\begin{equation}
\text{PSF(x,y)}=A *\text{exp}\left(-\left(\frac{x^2}{2\sigma_x^2}+\frac{y^2}{2\sigma_y^2}\right)\right)
\label{PSF_equ}
\end{equation}

Under ideal imaging conditions, when a target passes through a pixel, the projection of its two-dimensional point spread pattern along the motion direction approximates a one-dimensional Gaussian signal—naturally aligning with our modeling approach. Therefore, the proposed TPS mechanism not only captures the structural characteristics of transient target disturbances, but also possesses a solid physical foundation. Moreover, this mechanism eliminates the need for spatial annotations and makes no strong assumptions about the target’s shape, making it well-suited for weak target detection.

\subsubsection{Temporal Signal Modeling}
We provide the mathematical formulation of the pixel-level temporal signal and the TPS mechanism. Given an image sequence $S$, where each frame is denoted as $I$, the sequence can be expressed as:
\begin{equation}
S=\{I_1,I_2,I_3,\cdots, I_K\}
\label{seq_equ}
\end{equation}
where $K$ is the total number of frames in the sequence.

For a pixel located at the $m$-th row and $n$-th column of the sensor, its temporal intensity signal $s(k)$ is defined as:
\begin{equation}
s(k)=\{I_1(m,n),I_2(m,n),I_3(m,n),\cdots, I_K(m,n)\}
\label{ITS_equ}
\end{equation}

In practical surveillance systems, the imaging platform typically provides stabilized observations of a fixed scene. The temporal signal of each pixel is mainly affected by three components: the background intensity, detector noise, and potential target transient disturbance. Based on this, we can propose a binary hypothesis test as shown in equation~\ref{BHT_equ}, where $H_0$ denotes the hypothesis that the target has not passed and $H_1$ denotes the hypothesis that the target has passed.
\begin{align}
H_0: s(k)&=b(k)+n(k) \nonumber \\
H_1: s(k)&=b(k)+n(k)+t(k)
\label{BHT_equ}
\end{align}
where $b(k)$ represents the background signal, $n(k)$ denotes the noise, and $t(k)$ is the transient pulse signal introduced by a moving target.

Traditional methods often rely on specific statistical or structural assumptions about $b(k)$, $n(k)$, and $t(k)$ to design filters that suppress background and noise while enhancing the target component ~\cite{Niu_2024_Highb,Liu_2022_Moving}. However, such assumptions limit generalization and may fail when applied to scenarios that deviate from those assumptions. In contrast, our objective is to directly reconstruct $t(k)$ from $s(k)$ without relying on predefined models of background or noise.

Next, we introduce the proposed TPS mechanism, which generates a temporal Gaussian-shaped point signal as defined in Eq.~\ref{gk_equ}:

\begin{equation}
g(k)=A*\text{exp}(-\frac{(k-T)^2}{2\sigma_g^2})
\label{gk_equ}
\end{equation}
where, $A$ controls the amplitude of the pseudo-target signal, $T$ denotes the central time point of the target's appearance, and $\sigma_g$ determines the spread (i.e., temporal width) of the Gaussian distribution. According to the 3-$\sigma$ principle, approximately 99.74\% of the Gaussian signal's energy is concentrated within the interval $(T - 3\sigma_g, T + 3\sigma_g)$ , hence the effective temporal width can be approximated as $S \approx 6\sigma_g$. The parameters $A$, $T$, and $\sigma_g$ are randomly sampled to simulate diverse target characteristics in terms of intensity, duration, and speed.

When constructing the training dataset using the TPS mechanism, we embed the temporal Gaussian-shaped point signal under the null hypothesis $H_0$ (i.e., no real target present), resulting in the following observation model:
\begin{equation}
s(k)=b(k)+n(k)+g(k)
\label{TPS_equ}
\end{equation}

In practice, however, whether a pixel belongs to 
$H_0$ or $H_1$ (i.e., target present) is typically unknown, which is precisely the challenge that target detection seeks to address. As such, it is inevitable that some point signals $g(k)$ may be added to pixels already containing true target signals $t(k)$. Fortunately, this does not undermine the training process. Given the sparsity of real targets, the probability of pseudo-targets overlapping with true targets is extremely low. Moreover, our signal reconstruction network is capable of recovering multiple transient perturbations within a single temporal signal, ensuring the robustness of the supervision.

To train the network to reconstruct $t(k)$ from observed signals $s(k)$, we introduce $g(k)$ with randomly sampled parameters during data construction. Meanwhile, to ensure a consistent target representation at the output layer, the corresponding ground-truth supervision labels are normalized by setting $A=1$, regardless of the actual amplitude used during signal synthesis.

\subsection{Temporal Signal Reconstruction Network}
In this section, we present the TSRNet based on an encoder-decoder architecture. We first introduce the network architecture in detail, followed by a description of the weighted loss function specifically designed to address the severe class imbalance encountered during training.

\subsubsection{Network Architecture}
\begin{figure*}[!t]
	\centering
	\includegraphics[width=0.9\linewidth]{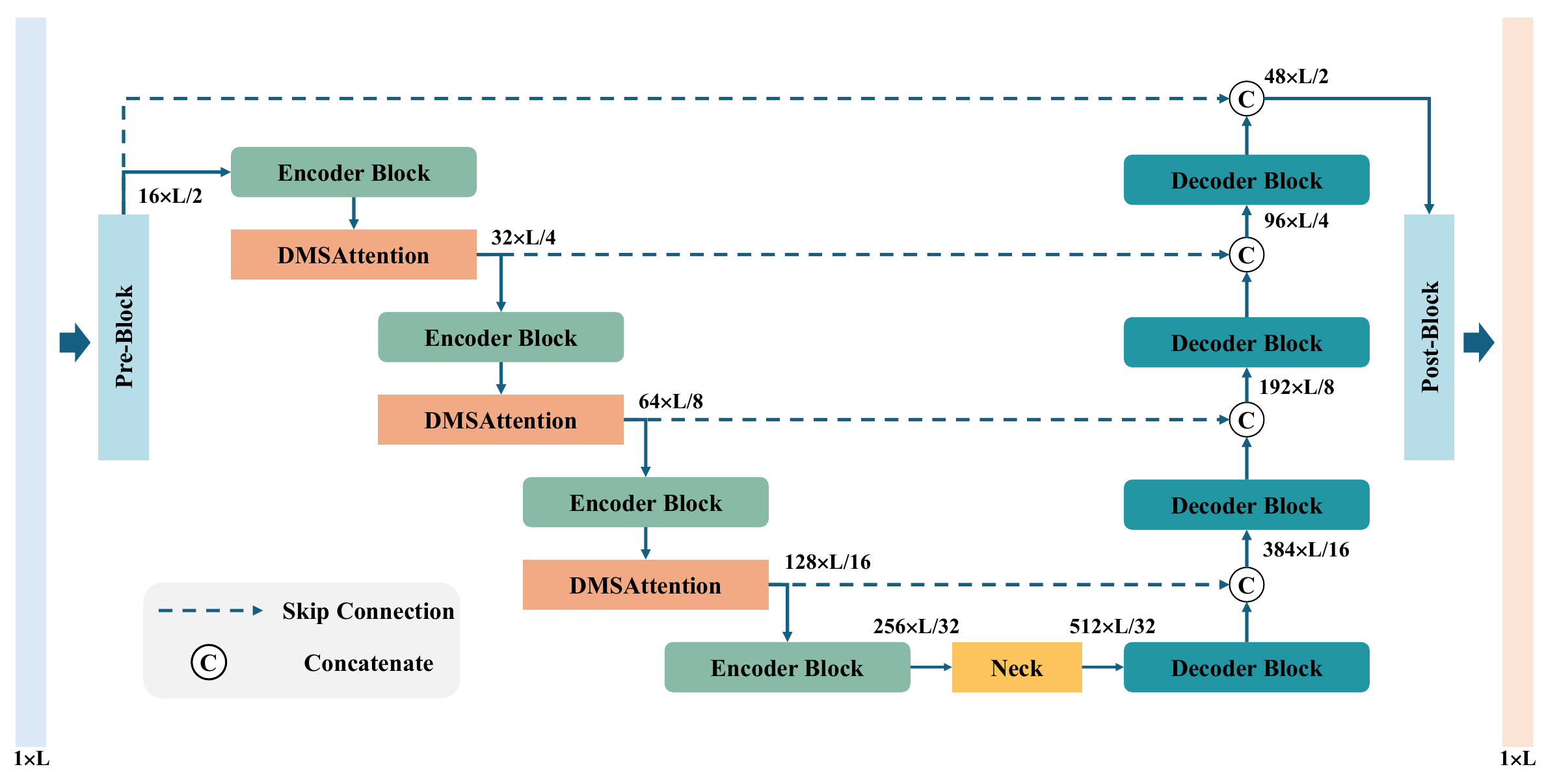}
	\caption{The architecture of the proposed TSRNet. The network adopts an encoder–decoder structure with dynamic multi-scale attention (DMSAttention) modules inserted after each encoder stage. The input 1D signal is first processed by a pre-block and then encoded through a series of encoder blocks, progressively downsampling the resolution while increasing channel dimensions. Skip connections deliver features to corresponding decoder blocks via concatenation, facilitating information flow. The final output is reconstructed through a post-block, restoring the original resolution. Dimensions are annotated as channels × length.}
	\label{network_fig}
\end{figure*}

As illustrated in Fig.~\ref{network_fig}, the proposed network adopts an encoder–decoder architecture to reconstruct the transient target signal $t(k)$ from the observed input $s(k)$. The encoder progressively downsamples the temporal input, extracting multi-scale features, while the decoder performs upsampling and signal restoration. Skip connections are introduced to preserve essential low-level characteristics such as shape and scale, and to enhance information flow across layers, which is critical for recovering weak and short-duration signals.

To further improve the network’s ability to focus on local perturbations, DMSAttention modules are embedded at multiple stages. These modules adaptively aggregate features from different temporal receptive fields, enhancing target-related activations while suppressing background noise. This design significantly improves the network’s robustness and sensitivity in low SNR scenarios. 

In addition, the network is designed to be lightweight and efficient, with only 1.84M parameters in total, making it well suited for deployment in real-time or resource-constrained environments. Despite its compact size, the architecture maintains strong feature representation capability through multi-scale encoding, attention-based enhancement, and skip-connected decoding.

Next, we describe the modules in the network in detail.

\begin{figure}[!t]
	\centering
	\includegraphics[width=\columnwidth]{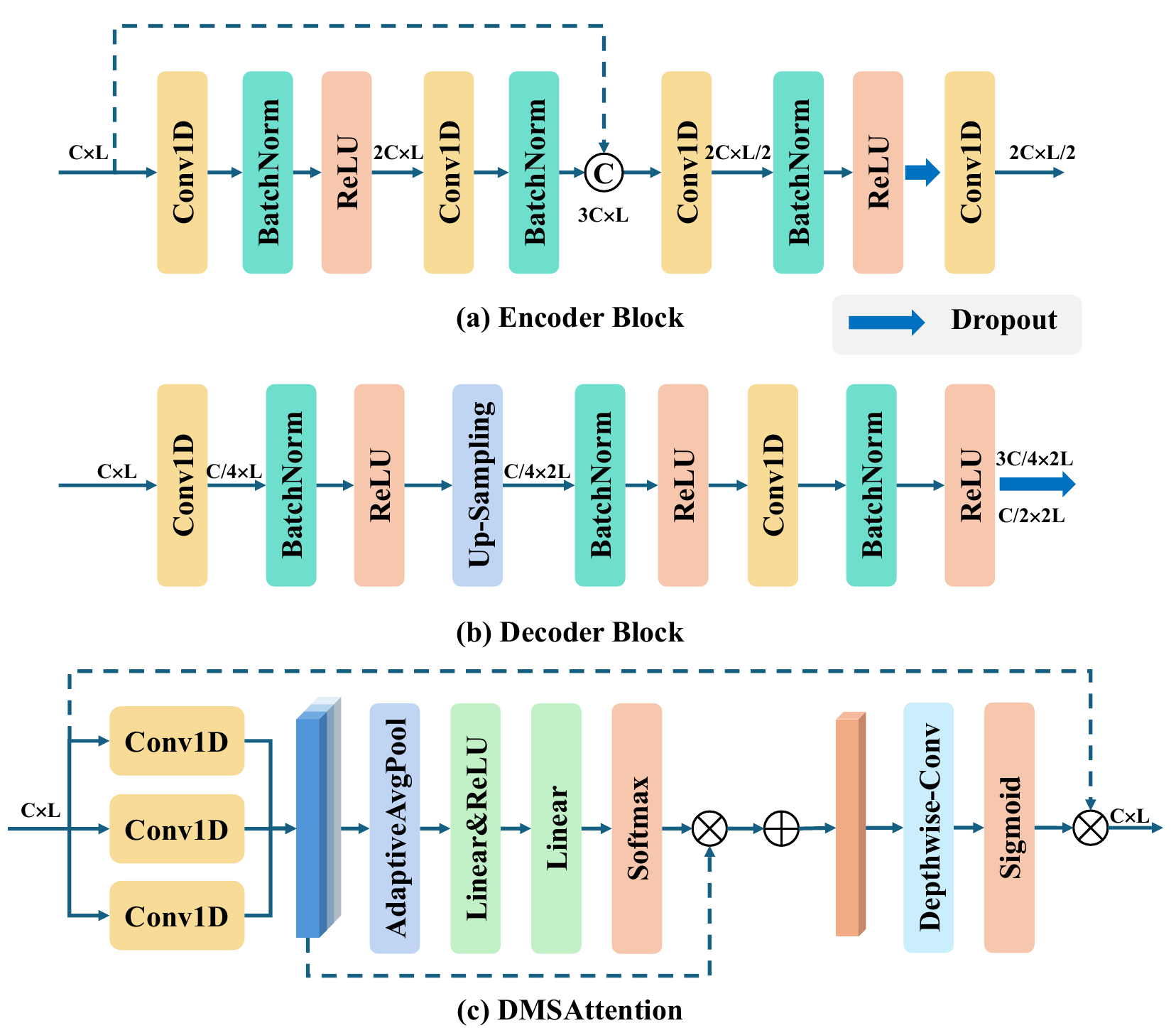}
	\caption{Architectures of key modules in the proposed TSRNet.
(a) Encoder Block progressively extracts temporal features through stacked Conv1D and BatchNorm layers, with skip connections and concatenation to retain original signal characteristics. A downsampling step is applied via strided convolution.
(b) Decoder Block restores temporal resolution through upsampling, followed by convolution and normalization layers to refine the reconstructed signal. Residual connections are used for multi-scale feature integration.
(c) DMSAttention module enhances target-related temporal perturbations by combining multi-scale depthwise convolutions with adaptive weight selection. The learned weights dynamically fuse features across different receptive fields, followed by a depthwise Conv1D and sigmoid gating to generate an attention mask.}
	\label{coder_fig}
\end{figure}

\begin{enumerate}

\item \textbf{Pre-Block}: The Pre-Block consists of a $7\times1$ 1D convolution followed by Batch Normalization and ReLU activation, aiming to enhance local feature perception. It then applies a max-pooling operation for temporal downsampling, which enlarges the receptive field and reduces computational complexity. Subsequently, a 1D convolution is used to increase the channel dimension, and a dropout layer is added to prevent overfitting.

\item \textbf{Encoder Block}: As illustrated in Fig.~\ref{coder_fig}(a), the Encoder Block is composed of two Conv1D layers with Batch Normalization to extract transient features of the target signal. A skip connection is introduced to concatenate shallow features with the current layer’s output, facilitating gradient propagation and preserving fine-grained details. Downsampling is performed using a stride-2 Conv1D, followed by BatchNorm and ReLU to refine the feature representation. Dropout is applied afterward to mitigate overfitting, and a final Conv1D layer is used to further enhance feature expression.

\item \textbf{DMSAttention}:As shown in Fig.~\ref{coder_fig}(c), due to variations in target velocity, imaging range, and pixel resolution, the temporal extent of target signals can vary significantly. To adaptively enhance target responses across multiple temporal scales, we propose a Dynamic Multi-Scale Attention module. Let the input feature be denoted as $ \mathbf{X} \in \mathbb{R}^{B \times C \times L} $, where $B$ is the batch size, $C$ is the number of channels, and $L$ is the temporal fetaure length. We first extract multi-scale local features using 1D convolutions with different kernel sizes $k_i \in \{5, 11, 21\}$. Each scale $i$ produces a feature map:
\begin{equation}
   \mathbf{F}_i = \text{Conv1D}_{k_i}(\mathbf{X}) \in \mathbb{R}^{B \times C \times L} 
   \label{F_i_equ}
\end{equation}

These are then stacked across a new dimension:
\begin{equation}
\mathbf{F} = \text{Stack}(\mathbf{F}_1, \mathbf{F}_2, \mathbf{F}_3) \in \mathbb{R}^{B \times 3 \times C \times L}
   \label{stack_equ}
\end{equation}

To compute the dynamic weights across scales, we first apply global adaptive average pooling (GAP) to compress temporal information:
\begin{equation}
\mathbf{z} = \text{GAP}(\mathbf{X}) \in \mathbb{R}^{B \times C}
   \label{avgpool_equ}
\end{equation}
This vector is passed through a two-layer fully connected network with softmax activation to produce per-scale weights:
\begin{equation}
\mathbf{w} = \text{Softmax}\left( \mathbf{W}_2 \cdot \sigma\left( \mathbf{W}_1 \cdot \mathbf{z} \right) \right) \in \mathbb{R}^{B \times 3}
   \label{softmax_equ}
\end{equation}
The weights are then reshaped and applied to the stacked features for weighted fusion:
\begin{equation}
\mathbf{F}_{\text{fused}} = \sum_{i=1}^{3} w_i \cdot \mathbf{F}_i \in \mathbb{R}^{B \times C \times L}
   \label{fused_equ}
\end{equation}

Next, channel-wise attention is applied to emphasize informative temporal features. A lightweight channel attention mechanism is used:
\begin{equation}
\mathbf{A}_{\text{channel}} = \sigma \left( \mathbf{Conv1D} (\delta\left( \mathbf{Conv1D}( \mathbf{F}_{\text{fused}}) \right)) \right) \in \mathbb{R}^{B \times C \times L}
   \label{channel_equ}
\end{equation}
Finally, the attention mask is applied back to the input feature via element-wise multiplication $\mathbf{X}' = \mathbf{X} \odot \mathbf{A}_{\text{channel}}$. The overall DMSAttention module enhances temporal local perturbations adaptively while maintaining lightweight computation.

\item \textbf{Neck}: Serving as a bridge between the encoder and decoder, the Neck module employs a $3\times1$ convolution to expand the feature dimension to 512 channels. This provides a richer semantic representation and enhances global context for subsequent signal reconstruction.

\item \textbf{Decoder Block}: As shown in Fig.~\ref{coder_fig}(b), the Decoder Block consists of two Conv1D layers with BatchNorm and ReLU activation to refine the feature map. Temporal upsampling is achieved through ConvTranspose1D. At each decoding stage, the upsampled features are concatenated with the corresponding encoder features to restore temporal resolution and preserve target details. Dropout is also applied in each decoder block to improve generalization.

\item \textbf{Post-Block}: The final output undergoes an upsampling operation followed by two 1D convolutions to produce a single-channel output. A sigmoid activation function maps the result to the range $(0,1)$, representing the probability of target presence at each time step, which serves as the final reconstructed signal.

\end{enumerate}

\subsubsection{Loss Function Design}

In target signal reconstruction, the data naturally exhibit significant class imbalance: most time steps correspond to background signals, while only a few contain target pulses.  
This creates a dual challenge—enhancing sparse target pulses while suppressing noisy background fluctuations.

To address this, we design a regression-based weighted loss function that emphasizes target regions and selectively penalizes false responses in the background. The total loss is defined as:

\begin{equation}
\mathcal{L}_{\text{total}} = \alpha \cdot \mathcal{L}_{\text{target}} + \beta \cdot \mathcal{L}_{\text{background}}
\label{total_loss_equ}
\end{equation}

Here, $\alpha$ and $\beta$ control the relative importance of target enhancement and background suppression.  
We define a binary mask based on the Gaussian point label $y(t)$:

\begin{equation}
\text{MASK}(t) = 
\begin{cases} 
1, & y(t) > 0  \\
0, & y(t) = 0
\end{cases}
\label{mask_def}
\end{equation}

The target and background losses are computed as:

\begin{equation}
\mathcal{L}_{\text{target}} = \frac{1}{\sum{\operatorname{MASK}}} \sum_{t:\operatorname{MASK}(t)=1} \left( \hat{y}(t) - y(t) \right)^2
\label{loss_target_equ}
\end{equation}

\begin{equation}
\mathcal{L}_{\text{background}} = \frac{1}{\sum{(1 - \operatorname{MASK})}} \sum_{t:\operatorname{MASK}(t)=0} \max\left( 0, \hat{y}(t) - \delta \right)^2
\label{loss_background_equ}
\end{equation}

Here, $\hat{y}(t)$ is the output of TSRNet, and $\delta$ is a suppression threshold applied to background predictions.  
Only when background responses exceed $\delta$ do they contribute to the loss, which helps reduce false positives while tolerating low-level fluctuations.

In practice, $\alpha$ and $\beta$ are selected based on signal-to-noise conditions. 
For weak targets and high background noise, a larger $\alpha$ and smaller $\beta$ are preferred.  
Since the target label is a Gaussian point signal with peak 1, we set $\delta$ to 5\%–10\% of the peak value (i.e., $\delta \in [0.05, 0.1]$) to balance false suppression and tolerance.

\subsection{Graph-based Trajectory Mining Algorithm}
\begin{figure}[t]
	\centering
	\includegraphics[width=\columnwidth]{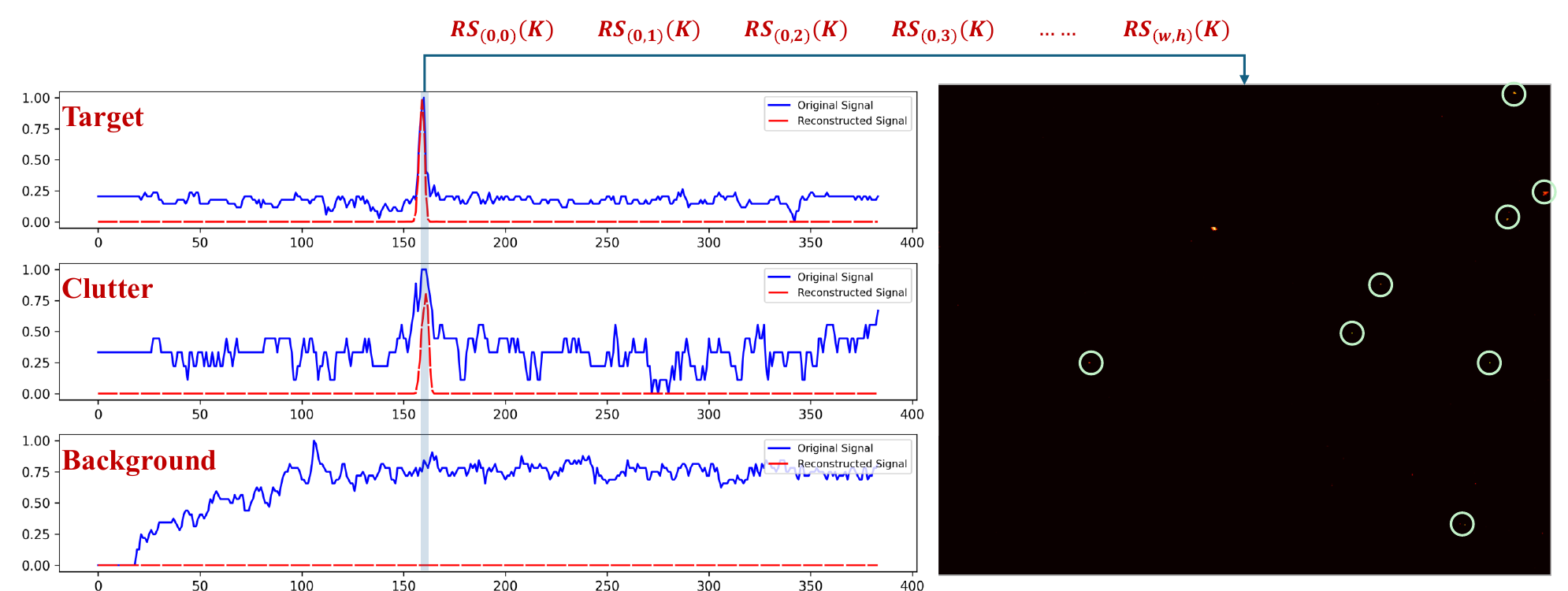}
	\caption{Illustration of TSRNet performance on different types of input signals and its impact on detection. On the left, clutter signals that closely resemble true targets are mistakenly enhanced by the TSRNet. As a result, the reconstructed image at frame $K$ (shown on the right) contains false alarms caused by these clutter responses. In the figure $RS_{(i,j)}$ denotes the reconstructed signal (RS) of pixel (i,j).}
	\label{false_alarm_fig}
\end{figure}

Although the proposed TSRNet effectively suppresses background signals and enhances target signals in most cases, it still retains certain clutter responses that closely resemble target-like pulses, especially under low SNR conditions. As illustrated in Fig.~\ref{false_alarm_fig}, such clutter often exhibits short-term intensity fluctuations similar to those of true targets, making them difficult to distinguish and resulting in false alarms.

To further improve the framework’s performance to differentiate true targets from clutter, we introduce a more discriminative representation from the spatiotemporal perspective. We observe that, in contrast to random noise and transient clutter, true targets generally exhibit consistent motion patterns and form continuous trajectories. Therefore, spatiotemporal continuity emerges as a key characteristic that fundamentally distinguishes real targets from clutter and serves as a reliable cue for refinement.

Motivated by this insight, we propose a Graph-based Trajectory Mining (GTM) algorithm to extract physically coherent target trajectories from the reconstructed signals. In this method, all detected points are formulated as nodes in a spatiotemporal graph. Edges are established between temporally adjacent and spatially close nodes to form candidate trajectories that exhibit consistent motion over time. This enables the effective suppression of isolated false positives caused by random noise. The full procedure is summarized in Algorithm~\ref{graph_mining}.

\begin{algorithm}[t]
\caption{Graph-based Trajectory Mining with Monte Carlo Optimization}
\label{graph_mining}
\KwIn{Set of detection points $P = \{(x_i, y_i, t_i)\}$, total frames $T$, number of trials $N$}
\KwOut{Final trajectory set $\mathcal{T}^*$}

Initialize best score $S^* \leftarrow -\infty$ \\
\For{$n \leftarrow 1$ \KwTo $N$}{
    Sample spatial distance threshold $d \sim \mathcal{U}(d_{min}, d_{max})$ \\
    Sample temporal gap $\Delta t \sim \mathcal{U}(\Delta t_{min}, \Delta t_{max})$ \\
    Sample minimum length $l \sim \mathcal{U}(l_{min}, l_{max})$ \\
    
    Construct edge list $E$ by connecting points within distance $d$ and temporal gap $\Delta t$ \\
    
    Build graph $G=(P, E)$ and extract connected components with length $\geq l$ \\
    Sort nodes in each component by time to form trajectories $\mathcal{T}$ \\
    
    Compute average length $\bar{L}$ and time coverage rate $C$ of trajectories \\
    Compute score $S = |\mathcal{T}| \cdot \bar{L} \cdot C$ \\
    
    \If{$S > S^*$}{
        $S^* \leftarrow S$ \\
        $(d^*, \Delta t^*, l^*) \leftarrow (d, \Delta t, l)$ \\
        $\mathcal{T}^* \leftarrow \mathcal{T}$
    }
}
\Return $\mathcal{T}^*$
\end{algorithm}

Our algorithm consists of three key components:
\begin{enumerate}
    \item \textbf{Graph Construction}: We begin by selecting all valid points$(x,y,t)$ from the reconstructed signals generated by TSRNet to form a point set $P$, Then, we construct the edge set $E$ based on spatiotemporal constraints. Specifically, an edge is established between two points $p_i, p_j$ if they are temporally adjacent ($|t_i - t_j| \leq \Delta t$) and spatially close ($|p_i - p_j| \leq d$).
        \begin{equation}
        E = \left\{ (i, j) \,\middle|\, \lVert p_i - p_j \rVert_2 \leq d,\ 0 < |t_i - t_j| \leq \Delta t \right\}
        \label{edge_equ}
        \end{equation}
    \item \textbf{Trajectory Extraction}: After constructing the graph, we identify connected components as potential target trajectories. Each connected subgraph is interpreted as a temporally ordered trajectory, representing continuous motion. To eliminate false trajectories caused by short-term fluctuations or isolated noisy points, we apply a minimum trajectory length threshold $l$ to filter out short segments.
    \item \textbf{Monte Carlo Optimization}: Due to significant variations in target velocity and clutter density across different scenes, manually tuning parameters—such as distance threshold $d$, maximum temporal gap $\Delta t$, and minimum trajectory length $l$—is challenging and often suboptimal. To address this, we adopt a Monte Carlo search strategy to automatically optimize these hyperparameters. Within a predefined number of random sampling iterations, we generate candidate parameter combinations and evaluate the quality of the extracted trajectories using a scoring function that jointly considers the number of trajectories, their average length, and temporal coverage.
\end{enumerate}

Through the proposed graph-based trajectory mining algorithm, we are able to effectively suppress false alarms and significantly improve detection performance. Importantly, our method does not require manual fine-tuning of sensitive hyperparameters. While parameters such as the number of Monte Carlo iterations $N$, temporal thresholds $\Delta t_{min}, \Delta t_{max}$, and trajectory length bounds $l_{min}, l_{max}$ are involved in Algorithm~\ref{graph_mining}, they are not sensitive and can be easily set based on the total number of frames $T$. Once the ranges are defined, the actual parameter selection is handled automatically through Monte Carlo sampling, eliminating the need for hand-crafted tuning across different scenarios.

\section{Experiments}

\subsection{Experiments Settings}
\subsubsection{Dataset}
The dataset used in our experiments is derived from SIRSTD ~\cite{Tong_2024_ST_Trans}, a large-scale infrared small target benchmark containing over 50,000 images, primarily focused on UAV targets. To construct a more challenging low-SNR evaluation benchmark, we selected 20 representative sequences from six distinct scenes in SIRSTD, based on scene stability and SNR levels. Within each scene, one sequence is used for training and the remaining sequences for testing, resulting in 6 training sequences and 14 testing sequences. This forms a training set of 1,453 frames and a testing set of 3191 frames in total.

Table~\ref{dataset_details_tab} provides a detailed summary of the selected sequences. Each scene is characterized by distinct background contexts, such as cloudy sky, dense forest, non-uniform and blurred backgrounds, which introduce different levels of noise and clutter. The testing sequences within each scene are designed to reflect SNR variations in the same environment, enabling robust performance evaluation under intra-scene perturbations. The average SNR across the entire dataset is 2.71, highlighting the low-SNR and weak signal conditions of our benchmark. In addition, our SNR was calculated using the formula ~\ref{SCR_equ}, and for the background region, we selected a $5\times5$ local background around the target.

\begin{equation}
\text{SNR}=\frac{|\mu_T-\mu_B|}{\sigma_B}
\label{SCR_equ}
\end{equation}

\begin{table}[htbp]
\caption{Details of the sequences in our low-SNR dataset. Each scene contains one training sequence and multiple testing sequences, covering various backgrounds and SNR levels.}
\label{dataset_details_tab}
\centering
\renewcommand\arraystretch{1.3}
\begin{tabular}{c c c c c >{\RaggedRight\arraybackslash}p{2cm}}
\toprule
Scene & Seq ID & Split & Frames & SNR & Description \\
\midrule
\multirow{4}{*}{1} 
 & S1-01 & Training & 399 & 2.85 & \multirow{4}{2cm}{Cloudy sky over rolling hills} \\
 & S1-02 & Testing  & 209 & 2.66 & \\
 & S1-03 & Testing  & 314 & 3.31 & \\
 & S1-04 & Testing  & 109 & 2.50 & \\
\midrule
\multirow{4}{*}{2} 
 & S2-01 & Training & 362 & 1.98 & \multirow{4}{2cm}{Dense forest} \\
 & S2-02 & Testing  & 111 & 0.84 & \\
 & S2-03 & Testing  & 337 & 1.92 & \\
 & S2-04 & Testing  & 171 & 0.47 & \\
\midrule
\multirow{3}{*}{3} 
 & S3-01 & Training & 98 & 4.65 & \multirow{3}{2cm}{Dark sky over rolling mountains} \\
 & S3-02 & Testing  & 81 & 4.14 & \\
 & S3-03 & Testing  & 262 & 4.60 & \\
\midrule
\multirow{4}{*}{4} 
 & S4-01 & Training & 133 & 6.10 & \multirow{4}{2cm}{Non-uniform blurred background} \\
 & S4-02 & Testing  & 85 & 5.36 & \\
 & S4-03 & Testing  & 127 & 6.37 & \\
 & S4-04 & Testing  & 124 & 5.51 & \\
\midrule
\multirow{2}{*}{5} 
 & S5-01 & Training & 200 & 2.79 & \multirow{2}{2cm}{Rolling hills with scattered trees} \\
 & S5-02 & Testing  & 869 & 2.51 & \\
\midrule
\multirow{3}{*}{6} 
 & S6-01 & Training & 261 & 0.66 & \multirow{3}{2cm}{Dense forest covering rolling hills} \\
 & S6-02 & Testing  & 264 & 0.86 & \\
 & S6-03 & Testing  & 128 & 0.88 & \\
\bottomrule
\end{tabular}
\end{table}

\subsubsection{Evaluation Metrics}
We adopt the 3D Receiver Operating Characteristic (3D ROC) metric ~\cite{Chang_2021_Effective}, which incorporates the detection probability $P_d$, false alarm rate $F_a$, and decision threshold $\tau$ as a comprehensive evaluation framework. The metrics $P_d$ and $F_a$ are calculated as follows:


\begin{align}
P_d = \frac{N_D}{N_T}, F_a = \frac{N_F}{N_B}
\label{pd_fa_equ}
\end{align}

Here, $P_d$ is a target-level metric that evaluates the detector's performance to correctly detect and locate targets (i.e., recall). Specifically, $N_D$ denotes the number of correctly detected true targets, and $N_T$ is the total number of ground-truth targets. $F_a$ is a pixel-level metric that measures the detector's performance to suppress false alarms, where $N_F$ is the number of background pixels incorrectly predicted as targets, and $N_B$ represents the number of actual background pixels.

Both $P_d$ and $F_a$ are calculated under a threshold $\tau \in [0,1]$. Varying $\tau$ generates a series of $(P_d, F_a)$ pairs, forming the traditional ROC curve. The Area Under the Curve (AUC) summarizes the performance across $\tau$ — higher AUC indicate better performance. However, conventional ROC analysis fails to fully capture a detector’s dual capabilities: enhancing weak targets and suppressing background clutter.

To address this, 3D ROC introduces $\tau$ into the ROC space, resulting in three projection curves: $(P_d, F_a)$, $(P_d, \tau)$, and $(F_a, \tau)$. The AUCs of these curves collectively reflect multiple aspects of detector performance. The following eight AUC-based metrics are used for comprehensive evaluation:
\begin{itemize}
    \item $\text{AUC}{(\text{D},\text{F})}$: Area under the $(P_d, F_a)$ curve, representing overall detection performance.
    \item $\text{AUC}{(\text{D},\tau)}$: Area under the $(P_d, \tau)$ curve, measuring detector sensitivity to varying thresholds.
    \item $\text{AUC}_{(\text{F},\tau)}$: Area under the $(F_a, \tau)$ curve, reflecting detector specificity.
    \item $\text{AUC}_{\text{TD}}$ (Target Detectability): Quantifies the ability to enhance target signals.
        \begin{equation}
            \text{AUC}_{\text{TD}} = \text{AUC}_{(\text{D},\text{F})} + \text{AUC}_{(\text{D},\tau)}
        \end{equation}
        
    \item $\text{AUC}_{\text{BS}}$ (Background Suppressibility): Reflects the capability to suppress background noise.
        \begin{equation}
            \text{AUC}_{\text{BS}} = \text{AUC}_{(\text{D},\text{F})} - \text{AUC}_{(\text{F},\tau)}
        \end{equation}
        
    \item $\text{AUC}_{\text{TD-BS}}$, $\text{AUC}_{\text{ODP}}$, and $\text{AUC}_{\text{SNPR}}$: Comprehensive indicators combining detection sensitivity and background suppression:
        \begin{equation}
            \text{AUC}_{\text{TD-BS}} = \text{AUC}_{(\text{D},\tau)} - \text{AUC}_{(\text{F},\tau)}
        \end{equation}
        \begin{equation}
            \text{AUC}_{\text{ODP}} = \text{AUC}_{(\text{D},\tau)} + (1 - \text{AUC}_{(\text{F},\tau)})
        \end{equation}
        \begin{equation}
            \text{AUC}_{\text{SNPR}} = \frac{\text{AUC}_{(\text{D},\tau)}}{\text{AUC}_{(\text{F},\tau)}}
\end{equation}
\end{itemize}

\subsubsection{Implementation Details}

During training, we extract pixel-wise temporal signals from the training sequences to construct the background signal pool. To improve efficiency, we do not extract training signals from every pixel; instead, we apply uniform subsampling with a stride of 4 for faster data collection. Gaussion point signals are then generated by adding short-duration temporal pulses with randomized amplitude 
$A$, temporal width $S$, and onset time $T$, to simulate varying target speeds and intensities. The range of randomness for $S$ and $A$ can be adapted based on the characteristics of different scenes (e.g., larger 
$S$ may be chosen for high-frame-rate sensors or slow-moving targets). In our experiments, the target amplitude $A$ is randomly sampled from the range $(10, 30)$, and the temporal width $S$ is sampled from the range $(5, 15)$. 

The network is optimized using the Adam optimizer, with a learning rate of $1\times10^{-3}$, batch size of 1000, and trained for 200 epochs. All deep learning-based experiments are conducted on a single NVIDIA RTX 4090 GPU, while traditional methods are implemented on an Intel i7-10700K CPU running at 3.8 GHz.

\subsection{Comparison to State-of-the-Arts Methods}
To validate the effectiveness and superiority of our proposed framework, we compare it against 12 state-of-the-art methods, including 5 traditional methods and 7 deep learning-based methods. Among the traditional methods, STRL-LBCM~\cite{10266665}, ASTTV-NTLA~\cite{9626011}, and RCTVW~\cite{liu2023representative} are multi-frame approaches, while FKRW~\cite{8705367} and PSTNN~\cite{PSTNN} are single-frame methods. The deep learning-based methods include ALCNet~\cite{9314219}, ACM~\cite{Dai_2021_WACV}, AGPCNet ~\cite{10024907}, DNANet ~\cite{li2022dense}, UIUNet~\cite{wu2022uiu}, SCTransNet~\cite{yuan2024sctransnet}, and DTUM~\cite{10321723}. Among them, DTUM is designed for multi-frame inputs, while the remaining models operate in a single-frame setting.

\subsubsection{Quantitative Evaluation}

\begin{figure*}[!t]
	\centering
	\includegraphics[width=\textwidth]{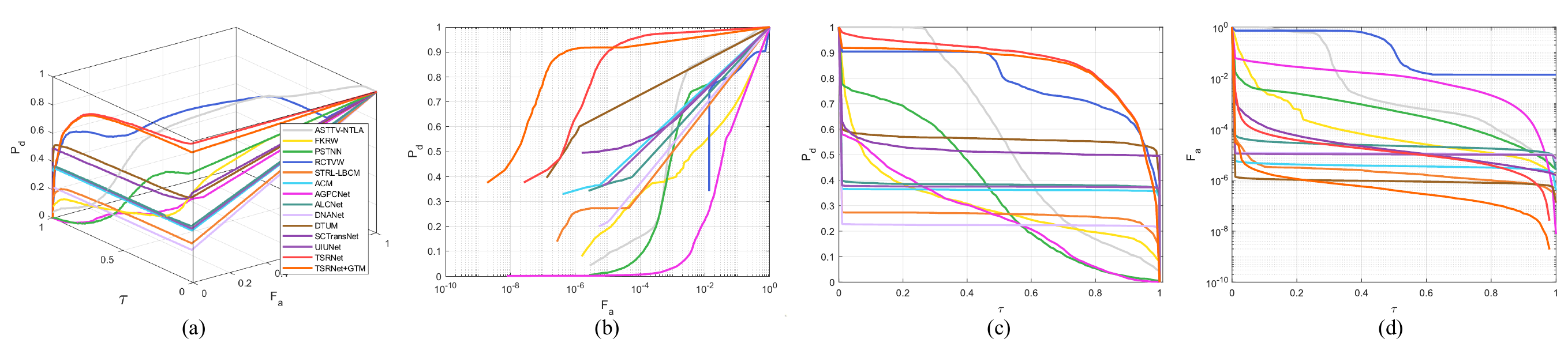}
	\caption{3D ROC curves and corresponding 2D ROC curves of comparison experiments. (a) 3D ROC curves. (b) 2D ROC curves of  $(P_d, F_a)$. (c) 2D ROC curves of  $(P_d, \tau)$. (d) 2D ROC curves of $(F_a, \tau)$.}
	\label{roc_fig}
\end{figure*}

As shown in Table~\ref{3droc_aucs_tab} and Fig.~\ref{roc_fig}, our framework (TSRNet and TSRNet+GTM) consistently outperforms existing mainstream traditional and deep learning-based methods across all 3D ROC-related metrics, demonstrating strong overall performance advantages. 

Our TSRNet achieves the best results in almost all metrics, including an AUC(D,F) of 0.9900, AUC(D,$\tau$) of 0.8584, AUC$_\text{TD}$ of 1.8484, AUC$_\text{TD-BS}$ of 0.8534, AUC$_\text{ODP}$ of 1.8534, and an overall score AUC$_\text{SNPR}$ reaching 169.6696. Compared with other deep learning-based methods such as SCTransNet (102.79) and DTUM (112.44), our model not only maintains a lower false alarm rate but also significantly improves the response to weak targets. This highlights the effectiveness of our framework in both suppressing background clutter and enhancing target signals. Compared with traditional multi-frame detection methods, such as ASTTV-NTLA (AUC(D,F) = 0.9779) and RCTVW (AUC(D,$\tau$) = 0.7949), our method exhibits clear advantages across all metrics.

Interestingly, when we integrate the GTM to form TSRNet+GTM, the performance on most quantitative metrics slightly decreases—for example, AUC(D,F) drops from 0.9900 to 0.9593, and AUC$_\text{TD-BS}$ decreases from 0.8534 to 0.8346. This is expected, due to the inherent trade-off between false alarm suppression and detection sensitivity in post-processing: the GTM module imposes stricter spatiotemporal connectivity constraints, which help eliminate isolated false positives but may also filter out some weak or fragmented target signals. This leads to a slight reduction in detection rate and, consequently, in some AUC scores.

Nevertheless, TSRNet+GTM exhibits a clear advantage in suppressing false alarms. As shown in Fig.~\ref{roc_fig}(b), after incorporating the GTM module, our method achieves a notably higher detection rate under extremely low false alarm rates compared to TSRNet without GTM. Specifically, as illustrated in Fig.~\ref{roc_fig}(d), the AUC(F,$\tau$) is reduced from 5.060E-3 to 5.001E-3. Although the numerical improvement appears modest, subsequent visual analyses reveal a substantial enhancement in result quality. This indicates that the GTM module strengthens the temporal coherence and trajectory continuity of detected targets, thereby improving interpretability and making the results more suitable for downstream tasks such as target tracking and behavior analysis.

In summary, TSRNet achieves the best overall performance across the majority of quantitative indicators, showcasing its strong capability in target enhancement and background suppression. Meanwhile, TSRNet+GTM, despite a slight trade-off in some metrics, provides superior interpretability and practicality by enhancing trajectory continuity and reducing false alarms—laying a solid foundation for subsequent high-level tasks.

\begin{table*}[h]
\caption{Quantitative comparison of 3D ROC metrics. Methods are grouped into traditional and deep-learning based. The best and second-best results are highlighted in \textbf{bold} and \underline{underlined}, respectively.}
\label{3droc_aucs_tab}
\centering
\resizebox{\textwidth}{!}{
\begin{tabular}{llcccccccc}
\toprule
\multicolumn{2}{c}{Method} & AUC(D,F) & AUC(D,$\tau$) & AUC(F,$\tau$) & AUC$_\text{TD}$ & AUC$_\text{BS}$ & AUC$_\text{TD-BS}$ & AUC$_\text{ODP}$ & AUC$_\text{SNPR}$ \\
\midrule
\multirow{5}{*}{\rotatebox[origin=c]{90}{\textbf{Traditional}}}
 & FKRW \cite{8705367}              & 0.8557 & 0.3045 & 12.217E-3 & 1.1601 & 0.8435 & 0.2923 & 1.2923 & 24.9215 \\
 & PSTNN \cite{PSTNN}         & 0.8855 & 0.3597 & 6.411E-3 & 1.2451 & 0.8790 & 0.3532 & 1.3532 & 56.1028 \\
 & ASTTV-NTLA \cite{9626011}       & \underline{0.9779} & 0.5746 & 230.870E-3 & 1.5525 & 0.7471 & 0.3437 & 1.3437 & 2.4888 \\
 & RCTVW \cite{liu2023representative} & 0.8876 & 0.7949 & 341.563E-3 & 1.6826 & 0.5461 & 0.4534 & 1.4534 & 2.3273 \\
 & STRL-LBCM \cite{10266665}       & 0.6368 & 0.2648 & 5.003E-3 & 0.9015 & 0.6318 & 0.2598 & 1.2598 & 52.9208 \\
\midrule
\multirow{9}{*}{\rotatebox[origin=c]{90}{\textbf{Deep-Learning}}}
 & ACM \cite{Dai_2021_WACV}        & 0.6826 & 0.3649 & 5.004E-3 & 1.0475 & 0.6776 & 0.3599 & 1.3599 & 72.9224 \\
 & AGPCNet \cite{10024907}         & 0.7660 & 0.2556 & 20.832E-3 & 1.0216 & 0.7452 & 0.2348 & 1.2348 & 12.2700 \\
 & ALCNet \cite{9314219}           & 0.6992 & 0.3856 & 5.023E-3 & 1.0849 & 0.6942 & 0.3806 & 1.3806 & 76.7707 \\
 & DNANet \cite{li2022dense}       & 0.6141 & 0.2273 & 5.011E-3 & 0.8413 & 0.6091 & 0.2222 & 1.2222 & 45.3507 \\
 & SCTransNet \cite{yuan2024sctransnet} & 0.8148 & 0.5202 & 5.061E-3 & 1.3350 & 0.8098 & 0.5151 & 1.5151 & 102.7874 \\
 & UIUNet \cite{wu2022uiu}         & 0.6896 & 0.3776 & 5.010E-3 & 1.0672 & 0.6846 & 0.3726 & 1.3726 & 75.3653 \\
 & DTUM \cite{10321723}            & 0.8004 & 0.5623 & \underline{5.001E-3} & 1.3627 & 0.7954 & 0.5573 & 1.5573 & 112.4405 \\
 & TSRNet+GTM (Ours)               & 0.9593 & \underline{0.8396} & \textbf{5.001E-3} & \underline{1.7989} & \underline{0.9543} & \underline{0.8346} & \underline{1.8346} & \underline{167.8747} \\
 & TSRNet (Ours)                   & \textbf{0.9900} & \textbf{0.8584} & 5.060E-3 & \textbf{1.8484} & \textbf{0.9849} & \textbf{0.8534} & \textbf{1.8534} & \textbf{169.6696} \\
\bottomrule
\end{tabular}
}
\end{table*}

\subsubsection{Visualization and Case Analysis}

\begin{figure*}[!t]
	\centering
	\includegraphics[width=\textwidth]{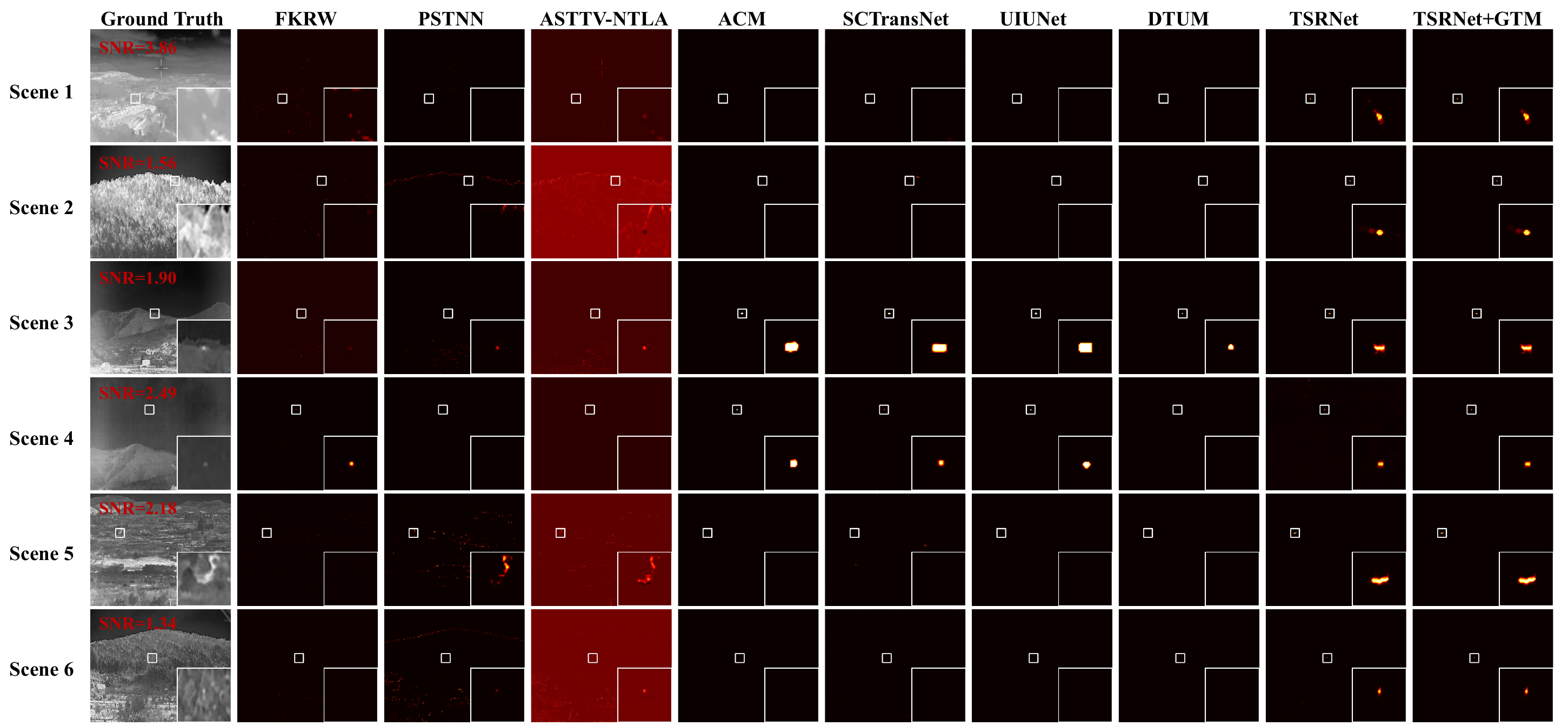}
	\caption{Visualization of representative single-frame detection results by different methods in Scenes 1–6.}
	\label{visual_frame_fig}
\end{figure*}

\begin{figure*}[!t]
	\centering
	\includegraphics[width=\textwidth]{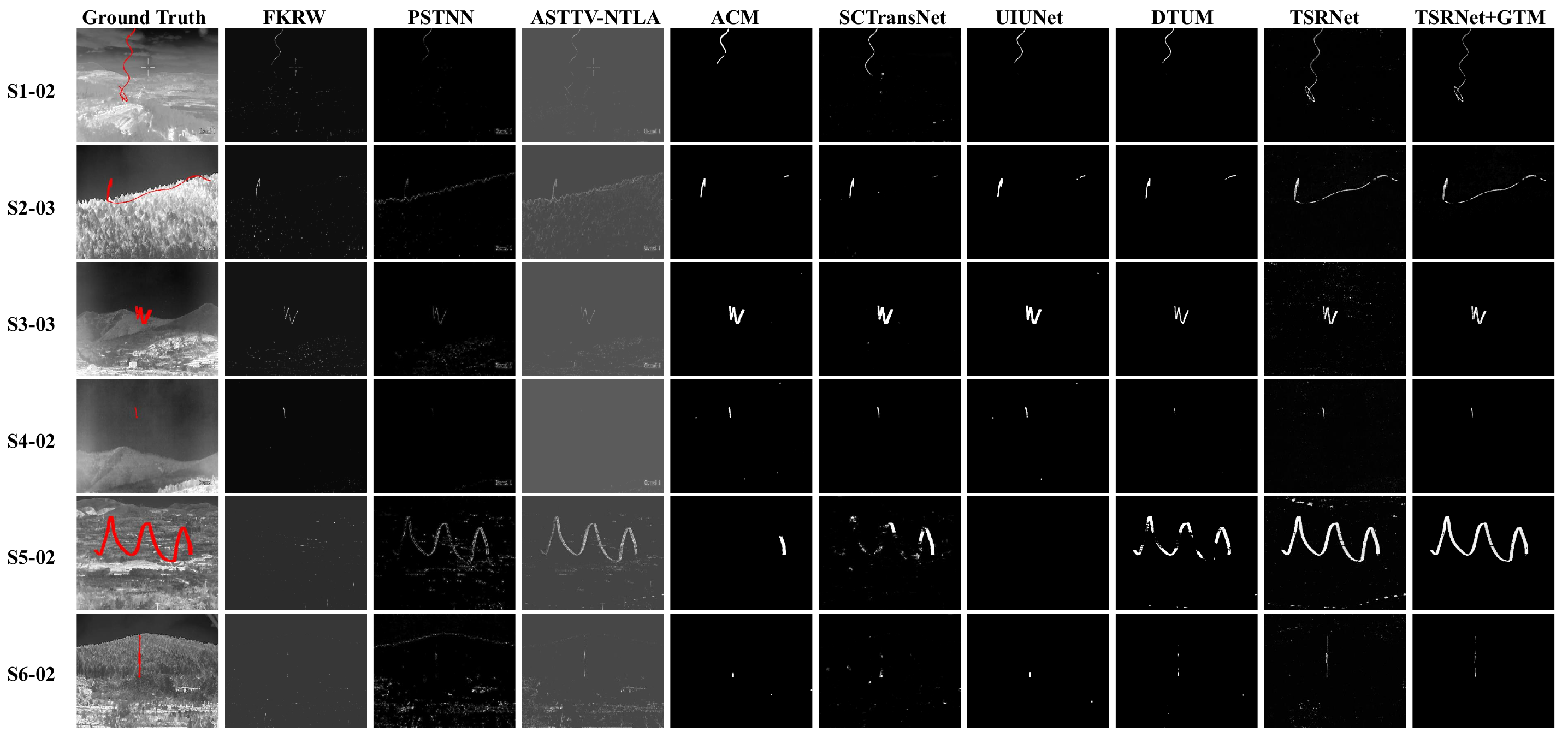}
	\caption{Visualization of detection trajectories over partial image sequences by different methods in Scenes 1–6.}
	\label{visual_trace_fig}
\end{figure*}

To qualitatively evaluate the detection performance of different methods, we present representative visualized results in Fig.~\ref{visual_frame_fig} and Fig.~\ref{visual_trace_fig}. Specifically, Fig.~\ref{visual_frame_fig} illustrates single-frame detection results produced by various methods. To more intuitively analyze each method's ability to suppress false alarms and detect targets, Fig.~\ref{visual_trace_fig} shows the resulting target motion trajectories.

As shown in Fig.~\ref{visual_frame_fig}, the selected frames exhibit low-SNR targets. Under these challenging conditions, our methods (TSRNet and TSRNet+GTM) successfully detect all targets with minimal false alarms, while most competing methods suffer from missed detections. Traditional methods tend to detect more targets but struggle with false alarm suppression, whereas deep learning-based methods are generally better at suppressing false alarms but often fail to detect weak targets. This highlights the superior capability of our framework in handling low-SNR target detection.

Trajectory visualizations in Fig.~\ref{visual_trace_fig} further emphasize this advantage. The trajectories detected by our methods closely match the ground truth, while those produced by other methods are often fragmented or incomplete. For example, most methods achieve satisfactory results in sequences like S3-03 and S4-02, where the SNR is relatively high, yielding continuous trajectories. However, in low-SNR sequences such as S1-02 and S2-03, their performance significantly degrades, revealing a lack of robustness to varying SNRs and background conditions. This observation supports our claim regarding the model's stronger generalization under low SNR conditions; a more detailed analysis of detection performance versus SNR will be presented later.

Furthermore, it is evident from the visualizations that the GTM module effectively eliminates false alarms. Compared to TSRNet alone, TSRNet+GTM produces cleaner and more consistent trajectories with no spurious detections—consistent with the conclusions drawn from our quantitative evaluations, and even more pronounced in these visual results.

\begin{figure*}[!t]
	\centering
	\includegraphics[width=\textwidth]{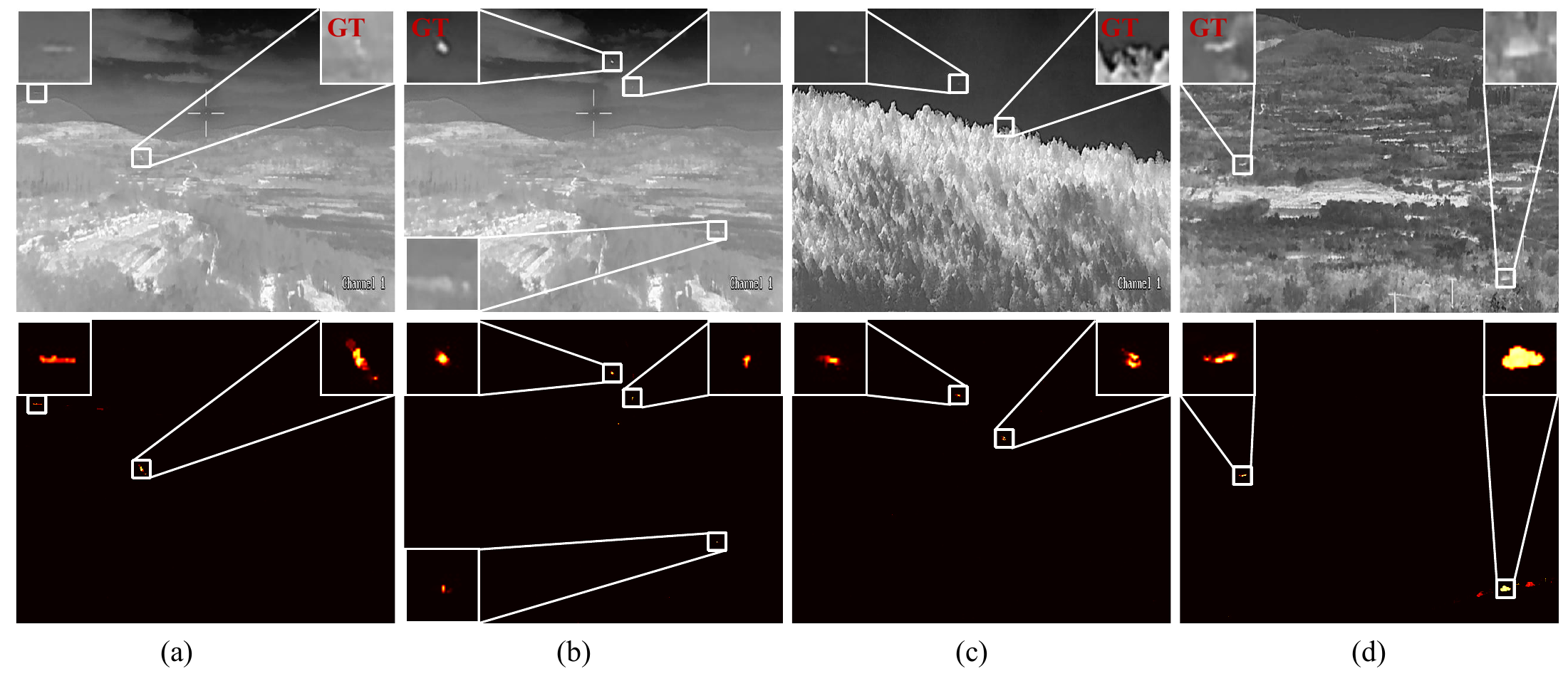}
	\caption{Visualisation results of all potential targets detected. GT (Ground Truth) in the figure refers to the pre-annotated targets and the rest of the targets are potential targets detected by our method.}
	\label{detect_bird_fig}
\end{figure*}

Moreover, we were pleasantly surprised to find that our detection method not only identifies the pre-annotated targets but also discovers additional targets that are not present in the ground-truth annotations, as illustrated in Fig.~\ref{detect_bird_fig}. Most of these additional detections appear to be birds flying through the scene. We consider this a meaningful outcome, as it demonstrates the effectiveness of our weakly supervised training strategy based on synthesized target-like pulse signals, which enables the model to detect as many potential targets as possible.

However, due to the high speed of some of these objects (e.g., gliding birds), many of the detections exhibit motion blur or ghosting artifacts, as shown in Fig.~\ref{detect_bird_fig}(d). Their trajectories are also less continuous, leading to their removal during the GTM post-processing. Nevertheless, this remains valuable—tasks such as bird detection in airport surveillance prioritize maximizing the number of detected objects in the scene.

\subsubsection{Analysis of Detection Performance under Different SNR}
\begin{figure}[h]
	\centering
	\includegraphics[width=\columnwidth]{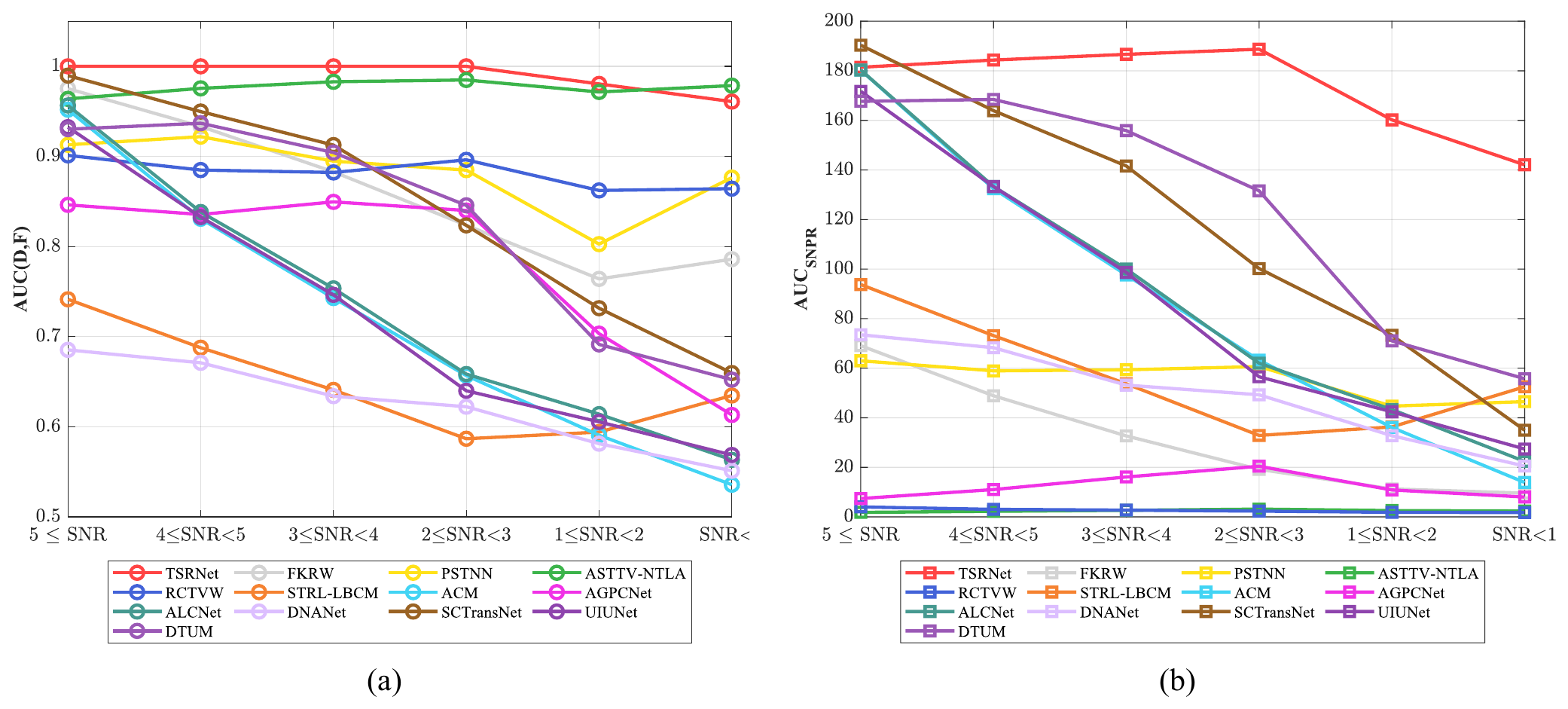}
	\caption{Trend plots of AUC(D,F) and AUC$_{\text{SNPR}}$ metrics under varying SNR conditions. (a) Variation trend of AUC(D,F) across different SNR. (b) Variation trend of AUC$_{\text{SNPR}}$ across different SNR.}
	\label{auc_snr_fig}
\end{figure}

To analyze the detection performance of different methods under varying SNRs, we divide the test set into six subsets based on target SNR: SNR < 1 (742 frames), 1$\le$SNR<2 (640 frames), 2$\le$SNR<3 (593 frames), 3$\le$SNR<4 (418 frames), 4$\le$SNR<5 (269 frames), and 5$\le$SNR (437 frames). Evaluation is performed on each subset using two comprehensive metrics: AUC(D,F) and AUC$_{\text{SNPR}}$. The experimental results are shown in Fig.\ref{auc_snr_fig} and Table\ref{tab:snr_comparison}.

As illustrated in the results, all methods exhibit performance degradation as SNR decreases. However, the degree of degradation varies significantly among different methods. Our proposed TSRNet consistently maintains high performance across all SNR ranges. Notably, under extremely low SNR conditions (SNR < 1), it still achieves an AUC(D,F) of 0.961 and an AUC$_{\text{SNPR}}$ of 142.05, which are significantly better than other competing methods. In contrast, methods such as FKRW, ACM, and UIUNet experience sharp performance drops in low SNR regimes (SNR < 3). For instance, FKRW's AUC$_{\text{SNPR}}$ drops below 10 under SNR < 1, indicating an almost complete failure to maintain valid detection.

It is worth noting that ASTTV-NTLA achieves a comparable AUC(D,F) score to TSRNet, particularly under SNR < 1, where it reaches 0.979. However, its AUC$_{\text{SNPR}}$ is only 2.37, indicating a serious issue with false alarm suppression and poor separability between weak signals and background interference. Therefore, the AUC(D,F) metric alone does not fully reflect the detection reliability under low SNR conditions. In contrast, AUC$_{\text{SNPR}}$ more effectively captures a model's ability to maintain detection precision amid signal noise.

Overall, TSRNet demonstrates excellent robustness and stability under low and extremely low SNR conditions, validating the effectiveness of our proposed framework for challenging weak target detection tasks.

\begin{table*}[h]
\centering
\caption{Comprehensive comparison of detection methods under varying SNR levels. Metrics include AUC(D,F) and AUC$_\text{SNPR}$ for each method across six SNR intervals. The best results are highlighted in \textbf{bold}.}
\label{tab:snr_comparison}
\resizebox{\textwidth}{!}{
\begin{tabular}{lcccccc}
\toprule
Method & SNR<1 & 1$\le$SNR<2 & 2$\le$SNR<3 & 3$\le$SNR<4 & 4$\le$SNR<5 & 5$\le$SNR \\
\midrule
TSRNet & 0.9609 / \textbf{142.05} & \textbf{0.9804} / \textbf{160.21} & \textbf{1.0000} / \textbf{188.72} & \textbf{1.0000} / \textbf{186.65} & \textbf{1.0000} / \textbf{184.35} & \textbf{1.0000} / \textbf{181.42} \\
FKRW \cite{8705367} & 0.7860 / 9.59 & 0.7641 / 11.28 & 0.8237 / 19.18 & 0.8829 / 32.67 & 0.9332 / 48.88 & 0.9753 / 69.09 \\
PSTNN \cite{PSTNN} & 0.8765 / 46.52 & 0.8027 / 44.70 & 0.8848 / 60.65 & 0.8948 / 59.36 & 0.9219 / 58.93 & 0.9129 / 62.99 \\
ASTTV-NTLA \cite{9626011} & \textbf{0.9786} / 2.37 & 0.9716 / 2.55 & 0.9848 / 3.14 & 0.9829 / 2.66 & 0.9755 / 2.25 & 0.9638 / 1.81 \\
RCTVW \cite{liu2023representative} & 0.8642 / 1.77 & 0.8622 / 1.85 & 0.8962 / 2.32 & 0.8822 / 2.73 & 0.8849 / 3.04 & 0.9012 / 4.02 \\
STRL-LBCM\cite{10266665} & 0.6347 / 52.51 & 0.5941 / 36.32 & 0.5867 / 32.87 & 0.6409 / 53.70 & 0.6877 / 73.14 & 0.7415 / 93.75 \\
ACM \cite{Dai_2021_WACV}\cite{Dai_2021_WACV} & 0.5357 / 13.90 & 0.5906 / 36.09 & 0.6568 / 63.22 & 0.7428 / 97.61 & 0.8309 / 132.41 & 0.9519 / 180.27 \\
AGPCNet\cite{10024907} & 0.6131 / 8.03 & 0.7033 / 10.87 & 0.8400 / 20.43 & 0.8496 / 16.11 & 0.8356 / 11.03 & 0.8462 / 7.41 \\
ALCNet\cite{9314219} & 0.5633 / 22.39 & 0.6140 / 43.33 & 0.6585 / 62.01 & 0.7536 / 100.01 & 0.8383 / 133.31 & 0.9565 / 180.31 \\
DNANet\cite{li2022dense} & 0.5512 / 20.52 & 0.5812 / 32.76 & 0.6223 / 49.27 & 0.6340 / 53.17 & 0.6710 / 68.26 & 0.6854 / 73.49 \\
SCTransNet \cite{yuan2024sctransnet} \cite{yuan2024sctransnet} & 0.6596 / 34.96 & 0.7316 / 73.29 & 0.8235 / 100.21 & 0.9126 / 141.58 & 0.9498 / 163.91 & 0.9897 / 190.33 \\
UIUNet \cite{wu2022uiu} & 0.5687 / 27.40 & 0.6055 / 42.39 & 0.6400 / 56.54 & 0.7464 / 98.62 & 0.8327 / 133.29 & 0.9325 / 171.73 \\
DTUM \cite{10321723} & 0.6523 / 55.75 & 0.6914 / 71.03 & 0.8457 / 131.58 & 0.9043 / 155.86 & 0.9368 / 168.43 & 0.9302 / 167.69  \\
\bottomrule
\end{tabular}
}
\end{table*}

\subsubsection{Computational Complexity Analysis}
Table~\ref{complexity_tab} presents the computational complexity and inference efficiency of different methods in terms of GFLOPS (Giga Floating-point Operations Per Second), number of parameters (Params), and FPS (Frames Per Second). The input data for each method follows its default configuration. Traditional methods are implemented in MATLAB, while deep learning methods are implemented using PyTorch.

Traditional detection methods typically do not rely on neural network training, so metrics like model parameters and GFLOPS are not directly comparable with deep learning methods. However, due to their reliance on complex operations such as matrix decomposition or tensor processing, their practical runtime efficiency is often limited. For instance, FKRW~\cite{8705367} and RCTVW~\cite{liu2023representative} achieve only 2.35 FPS and 5.64 FPS with 1-frame and 8-frame inputs, respectively. STRL-LBCM~\cite{10266665}, although capable of processing 16-frame sequences, runs at just 0.15 FPS, highlighting its limitation in real-time applications.

In contrast, deep learning methods benefit significantly from GPU acceleration. These models often strike a good balance between parameter size and inference speed. For example, ALCNet~\cite{9314219} requires only 1.89 GFLOPS and 0.43M parameters to achieve 212.42 FPS. Similarly, ACM~\cite{Dai_2021_WACV} delivers high real-time performance (182.59 FPS) with a compact model size (0.40M parameters). On the other hand, larger networks such as UIUNet~\cite{wu2022uiu} and AGPCNet~\cite{10024907} have significantly more parameters (50.54M and 12.36M, respectively), resulting in slower inference speeds of 36.27 FPS and 15.01 FPS, which are less suitable for real-time deployment. Notably, DTUM~\cite{10321723}, despite being a multi-frame method, achieves a decent inference speed of 54.04 FPS thanks to its minimal parameter count (0.30M) and moderate complexity (51.75 GFLOPS), outperforming some single-frame methods.

Our proposed TSRNet stands out with significant advantages in both computational efficiency and inference speed. It requires only 0.03 GFLOPS—the lowest among all compared methods—while achieving a remarkable inference speed of \textbf{1117.58} FPS, far exceeding other approaches. This is largely attributed to our efficient use of GPU parallelism: by transforming the image sequence into pixel-wise 1D temporal signals, we can process a large number of signals in parallel within a batch. Furthermore, the trajectory extraction module (GTM) employs a cKDTree-based acceleration strategy for graph construction. As a result, even with GTM included, TSRNet+GTM still achieves an impressive speed of \underline{671.79 FPS}, offering enhanced temporal consistency with only minimal additional computational cost.

In terms of model size, TSRNet contains just 1.84M parameters, which is significantly smaller than most deep networks. While DTUM~\cite{10321723} has the smallest parameter count (0.30M), its inference speed (54.04 FPS) still falls far short of TSRNet. Therefore, our method achieves a more favorable balance among parameter efficiency, runtime performance, and detection accuracy.

\begin{table*}[h]
\caption{Comparison of computational complexity and inference efficiency across different methods. The best results are highlighted in \textbf{bold}, and the second-best are \underline{underlined}.}
\label{complexity_tab}
\centering
\begin{tabular}{llccccccc}
\toprule
\multicolumn{2}{c}{Method} & Input Data & GFLOPS & Params & FPS \\
\midrule
\multirow{5}{*}{\rotatebox[origin=c]{90}{Traditional}}
 & FKRW \cite{8705367}              & 1 frame & - & - & 2.35 \\
 & PSTNN \cite{PSTNN}         & 1 frame & - & - & 1.67 \\
 & ASTTV-NTLA \cite{9626011}       & 3 frames & - & - & 0.09 \\
 & RCTVW \cite{liu2023representative} & 8 frames & - & - & 5.64 \\
 & STRL-LBCM \cite{10266665}       & 16 frames & - & - & 0.15 \\
\midrule
\multirow{9}{*}{\rotatebox[origin=c]{90}{Deep-Learning}}
 & ACM \cite{Dai_2021_WACV}        & 1 frame & 2.01 & \underline{0.40M} & 182.59 \\
 & AGPCNet \cite{10024907}         & 1 frame & 215.91 & 12.36M & 15.01 \\
 & ALCNet \cite{9314219}           & 1 frame & \underline{1.89} & 0.43M & 212.42 \\
 & DNANet \cite{li2022dense}       & 1 frame & 71.31 & 4.70M & 48.89 \\
 & SCTransNet \cite{yuan2024sctransnet} & 1 frame & 50.57 & 11.19M & 35.30 \\
 & UIUNet \cite{wu2022uiu}         & 1 frame & 272.13 & 50.54M & 36.27 \\
 & DTUM \cite{10321723}            & 5 frames & 51.75 & \textbf{0.30M} & 54.04 \\
 & TSRNet+GTM (Ours)               & 1D temporal signals & 0.03 & 1.84M & \underline{671.79} \\
 & TSRNet (Ours)                   & 1D temporal signals & \textbf{0.03} & 1.84M & \textbf{1117.58} \\
\bottomrule
\end{tabular}
\end{table*}

\subsection{Ablation Study}
To validate the effectiveness of the proposed modules, we conducted ablation studies. Given the simplicity of our network architecture, the ablation experiments focus primarily on the impact of the key DMSAttention module and the weighted loss function. We first constructed a baseline network (BaselineNet) by removing the DMSAttention module and compared it with the full TSRNet to assess the contribution of the attention mechanism to target enhancement. In addition, to demonstrate the advantage of our weighted loss function in addressing class imbalance, we replaced it with the standard MSELoss for comparison. The results are shown in Table~\ref{ablation_auc_tab}.

\begin{table}[htbp]
\caption{Ablation study on the impact of DMSAttention module and loss function design. TSRNet denotes the full model, BaselineNet is a simplified version without DMSAttention, and MSELoss replaces our designed loss.}
\label{ablation_auc_tab}
\centering
\renewcommand\arraystretch{1.2}
\resizebox{\columnwidth}{!}{
\begin{tabular}{lcccc}
\toprule
Network & TSRNet & TSRNet & BaselineNet & BaselineNet \\
Loss & Our Loss & MSE Loss & Our Loss & MSE Loss \\
\midrule
AUC(D,F)      & 0.9900 & 0.9784 & 0.9799 & 0.9343 \\
AUC(D,$\tau$) & 0.8584 & 0.7918 & 0.8139 & 0.6280 \\
AUC(F,$\tau$) & 0.0051 & 0.0050 & 0.0053 & 0.0051 \\
AUC$_{\text{TD}}$      & 1.8484 & 1.7701 & 1.7938 & 1.5623\\
AUC$_{\text{BS}}$      & 0.9849 & 0.9733 & 0.9746 & 0.9292\\
AUC$_{\text{TD-BS}}$   & 0.8534 & 0.7867 & 0.8087 & 0.6229\\
AUC$_{\text{ODP}}$     & 1.8534 & 1.7867 & 1.8087 & 1.6229\\
AUC$_{\text{SNPR}}$    & 169.67 & 156.85 & 154.44 & 123.8149 \\
\bottomrule
\end{tabular}
}
\end{table}

As shown in the results, TSRNet consistently outperforms BaselineNet across all evaluation metrics when both are trained with our proposed loss function. For instance, AUC(D, $\tau$) increases from 0.8139 to 0.8584, AUC$_{\text{TD-BS}}$ improves from 0.8087 to 0.8534, and AUC$_{\text{SNPR}}$ rises from 154.44 to 169.67. These improvements indicate that the DMSAttention module plays a crucial role in enhancing features of weak targets. When replacing our loss with MSELoss, most metrics drop significantly, suggesting that our weighted loss function handles class imbalance and suppresses background false alarms more effectively. The full TSRNet combined with the proposed loss function achieves the best performance across all metrics, demonstrating a strong synergy between the network architecture and the loss function in boosting detection under low SNR conditions.

\subsection{Sensitivity Analysis of Dataset Parameters}

\begin{figure*}[!t]
	\centering
	\includegraphics[width=\textwidth]{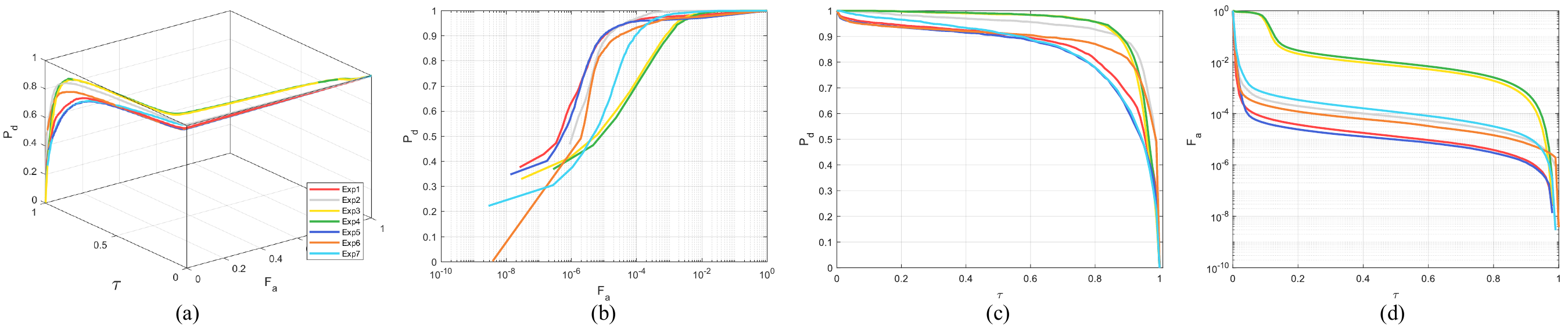}
	\caption{3D ROC curves and corresponding 2D ROC curves of parameter ranges experiments. (a) 3D ROC curves. (b) 2D ROC curves of  $(P_d, F_a)$. (c) 2D ROC curves of  $(P_d, \tau)$. (d) 2D ROC curves of $(F_a, \tau)$.}
	\label{param_roc_fig}
\end{figure*}

To evaluate the sensitivity of TSRNet to the distribution of pulse signal parameters in the training data, we conducted a series of experiments in which the network architecture and loss function were kept fixed, while varying the range of synthetic signal parameters used for training. The two key parameters under consideration are the pulse peak amplitude $A$ and the temporal width $S$. In the default setting (Exp1), the training signals are generated with $A \in(10,30)$ and $S\in(5,15)$. The background signals are extracted uniformly from the training set using a stride of 4, and are consistent across all experiments. The experimental results are shown in Fig.\ref{param_roc_fig} and Table\ref{param_robustness_tab}.

The results demonstrate that different parameter settings have varying degrees of impact on detection performance. Overall, changes in the temporal width $S$ tend to have a smaller effect, whereas variations in the amplitude $A$ show greater sensitivity. The detailed analysis is as follows:

\begin{itemize}
    \item Effect of varying $S$: In Exp2, the width range is expanded to $S\in(1,30)$, leading to improved performance across multiple metrics compared to the default setting. This suggests that incorporating a wider range of pulse durations—particularly shorter pulses—enables the network to learn more generalizable temporal patterns. This observation is further supported by Exp6 and Exp7: Exp6 narrows the width range to $S\in(1,10)$ while still achieving strong performance, whereas Exp7 shifts the range upward to $S\in(10,20)$ and results in a noticeable drop in performance. These results imply that excluding short pulses from the training distribution may limit the model’s ability to detect weak targets. Therefore, the choice of $S$ is relatively flexible, and including shorter pulse durations is generally beneficial.
    \item Effect of varying $A$: In contrast, the network is more sensitive to the choice of amplitude range. Experiments Exp3, Exp4, and Exp5 adjust the upper and lower bounds of $A$, with mixed results. Although target enhancement metrics such as AUC(D,$\tau$) remain stable, the false alarm suppression metric AUC(F,$\tau$) increases by an order of magnitude in Exp3 and Exp4. This indicates that incorporating too many low-amplitude signals (e.g., $A\in(1,20)$) introduces excessive noise during training, reducing the separability between target and background signals and ultimately degrading detection performance.
\end{itemize}

Notably, although the default configuration (Exp1) does not achieve the best overall performance, it yields the best result in false alarm control, as reflected by the lowest AUC(F,$\tau$). Furthermore, it confirms the effectiveness of our weakly supervised framework, which achieves strong performance without requiring elaborate parameter tuning.

\begin{table*}[h]
\caption{Ablation study on the robustness of training parameter ranges. All models use the same TSRNet architecture but are trained using different ranges of synthetic signal amplitudes and widths. The default setting (Exp1) corresponds to $A \in (10,30)$ and $S \in (5,15)$. The best and second-best results are highlighted in \textbf{bold} and \underline{underlined}, respectively.}
\label{param_robustness_tab}
\centering
\resizebox{\textwidth}{!}{
\begin{tabular}{clcccccccc}
\toprule
Experiment & Parameter Ranges & AUC(D,F) & AUC(D,$\tau$) & AUC(F,$\tau$) & AUC$_\text{TD}$ & AUC$_\text{BS}$ & AUC$_\text{TD-BS}$ & AUC$_\text{ODP}$ & AUC$_\text{SNPR}$ \\
\midrule
Exp 1 & $A \in (10,30)$, $S \in (5,15)$     & 0.9900 & 0.8584 & \textbf{5.060E-3} & 1.8484 & 0.9849 & 0.8534 & 1.8534 & 169.6696 \\
Exp 2 & $A \in (10,30)$, $S \in (1,30)$ & \textbf{1.0000} & 0.9357 & 5.345E-3 & 1.9357 & \textbf{0.9946} & \textbf{0.9304} & \textbf{1.9304} & \textbf{175.0487} \\
Exp 3 & $A \in (1,40)$, $S \in (5,15)$  & 0.9995 & \underline{0.9376} & 9.326E-2 & \underline{1.9371} & 0.9062 & 0.8443 & 1.8443 & 10.0535 \\
Exp 4 & $A \in (1,20)$, $S \in (5,15)$    & 0.9993 & \textbf{0.9434} & 9.970E-2 & \textbf{1.9427} & 0.8996 & 0.8437 & 1.8437 & 9.4622 \\
Exp 5 & $A \in (20,40)$, $S \in (5,15)$     & 0.9859 & 0.8396 & \underline{5.146E-3} & 1.8255 & 0.9808 & 0.8345 & 1.8345 & 163.1530 \\
Exp 6 & $A \in (10,30)$, $S \in (1,10)$  & 0.9882 & 0.8857 & 5.188E-3 & 1.8739 & 0.9830 & \underline{0.8805} & \underline{1.8805} & \underline{170.7338} \\
Exp 7 & $A \in (10,30)$, $S \in (10,20)$  & \underline{0.9997} & 0.8534 & 5.644E-3 & 1.8531 & \underline{0.9940} & 0.8478 & 1.8478 & 151.1988 \\
\bottomrule
\end{tabular}
}
\end{table*}

\section{Discussion}

In this section, we discuss the strengths and limitations of our proposed framework.

Traditional detection paradigms typically aim to localize the target’s spatial coordinates $(x,y)$ at each given frame $t$. In contrast, our framework inverts this perspective by focusing on each spatial location $(x,y)$ and determining the temporal moments $t$ at which a target appears. This inversion of perspective forms the foundational motivation of our work.

Our core innovation lies in the design of the Temporal Point Supervision (TPS) mechanism, which generates labels for training without any manual annotation. This not only eliminates the need for human labeling but also significantly reduces the cost of data preparation and model training.

We designed TSRNet, a lightweight encoder-decoder architecture tailored for signal reconstruction at the pixel level. It incorporates a Dynamic Multi-Scale Attention (DMSAttention) module to selectively enhance weak target responses, whose effectiveness has been validated through extensive ablation studies. Furthermore, the efficiency of our TSRNet is noteworthy — with GPU acceleration, it achieves over 1000 FPS, making it highly suitable for real-time applications. On the constructed low-SNR dataset, our method consistently outperforms a wide range of state-of-the-art baselines across all major metrics.

To mitigate false alarms caused by clutter responses, we propose a Monte Carlo optimized graph-based trajectory mining algorithm (GTM), which automates parameter tuning during inference and improves overall detection robustness.

Fundamentally, our framework functions as a target pulse signal reconstruction pipeline, and its applicability is not limited to weak target detection. It can be extended to other signal reconstruction tasks such as event-based vision and neuronal spike train analysis.

Despite these advantages, our framework has several limitations. First, the method assumes a relatively stable scene — not necessarily a static background, but one where temporal signal consistency exists. In highly dynamic environments with rapid background changes, the temporal distinction between target and background signals becomes blurred, compromising detection performance. Second, the framework is inherently designed to detect moving targets relative to the background. Stationary objects, which do not produce pulse-like signatures in the temporal domain, are beyond the scope of this framework. Lastly, our current graph-based trajectory mining strategy relies on relatively simple filtering rules, which may inadvertently discard valid target trajectories. This aspect warrants further refinement. And we plan to address these limitations in the future work.

\section{Conclusion}

This paper presents a novel weakly supervised framework for low-SNR moving target detection based on temporal point supervision, effectively addressing the challenge of low SNR detection without requiring any manual annotations. We reformulate the detection unit from conventional image frames to pixel-level temporal signals, leveraging the transient pulse characteristics induced by target motion to generate temporal point-labels. We then introduce TSRNet, a lightweight temporal signal reconstruction network and a weighted loss function. Under extremely low SNR conditions, TSRNet achieves best detection performance. With only 1.84M parameters and inference speeds exceeding 1000 FPS, it offers strong potential for real-word surveillance and early warning applications. To address false positives and associate detections over time, we propose a graph-based trajectory mining algorithm (GTM), which incorporates Monte Carlo-based parameter optimization for robust, automatic trajectory inference. Extensive experiments demonstrate clear advantages over existing methods, and ablation studies confirm the contributions of each component. Despite its strengths, the method assumes moving targets and stable backgrounds, limiting its use in dynamic or static-target scenarios. Future work will focus on extending applicability to more complex environments.

\printcredits

\bibliographystyle{cas-model2-names}

\bibliography{cas-refs}

\end{document}